\documentclass[runningheads]{llncs}
\usepackage{graphicx}

\usepackage{tikz}
\usepackage{comment} 
\usepackage{amsmath,amssymb} 
\usepackage{color}

\usepackage{booktabs}
\usepackage{xspace}
\usepackage[title]{appendix}
\usepackage{placeins}
\usepackage[pagebackref=true,breaklinks=true,letterpaper=true,colorlinks,bookmarks=false]{hyperref}
\usepackage{lipsum}

\newcommand\blfootnote[1]{%
  \begingroup
  \renewcommand\thefootnote{}\footnote{#1}%
  \addtocounter{footnote}{-1}%
  \endgroup
}

\makeatletter
\DeclareRobustCommand\onedot{\futurelet\@let@token\@onedot}
\def\@onedot{\ifx\@let@token.\else.\null\fi\xspace}

\def\eg{\emph{e.g}\onedot} 
\def\ie{\emph{i.e}\onedot} 
 
 \def\vs{\emph{vs}\onedot}
 
\def\etal{\emph{et al}\onedot}

\newcommand{\figref}[1]{Fig\onedot~\ref{#1}}
\newcommand{\equref}[1]{Eq\onedot~\eqref{#1}}

\newcommand{\tabref}[1]{Tab\onedot~\ref{#1}}

\usepackage{pifont}
\newcommand{\cmark}{\ding{51}}

\begin{document}
\pagestyle{headings}
\mainmatter
\def\ECCVSubNumber{1564}  

\title{Axial-DeepLab: Stand-Alone Axial-Attention for Panoptic Segmentation} 


\titlerunning{Axial-DeepLab}
%
\author{
Huiyu Wang\inst{1}\thanks{Work done while an intern at Google.}
\and
Yukun Zhu\inst{2} \and
Bradley Green\inst{2} \and
Hartwig Adam\inst{2} \and
Alan Yuille\inst{1} \and
Liang-Chieh Chen\inst{2}}
\authorrunning{H. Wang et al.}
\institute{Johns Hopkins University \\
\and
Google Research
}
\maketitle

\begin{abstract}
Convolution exploits locality for efficiency at a cost of missing long range context. Self-attention has been adopted to augment CNNs with non-local interactions. Recent works prove it possible to stack self-attention layers to obtain a fully attentional network by restricting the attention to a local region. In this paper, we attempt to remove this constraint by factorizing 2D self-attention into two 1D self-attentions. This reduces computation complexity and allows performing attention within a larger or even global region. In companion, we also propose a position-sensitive self-attention design. Combining both yields our position-sensitive axial-attention layer, a novel building block that one could stack to form axial-attention models for image classification and dense prediction. We demonstrate the effectiveness of our model on four large-scale datasets. In particular, our model outperforms all existing stand-alone self-attention models on ImageNet. Our Axial-DeepLab\blfootnote{\url{https://github.com/csrhddlam/axial-deeplab}} improves 2.8\% PQ over {\it bottom-up} state-of-the-art on COCO test-dev. This previous state-of-the-art is attained by our small variant that is 3.8$\times$ parameter-efficient and 27$\times$ computation-efficient. Axial-DeepLab also achieves state-of-the-art results on Mapillary Vistas and Cityscapes.
\end{abstract}
\section{Introduction}

Convolution is a core building block in computer vision. Early algorithms employ convolutional filters to blur images, extract edges, or detect features. It has been heavily exploited in modern neural networks \cite{lecun1998gradient,krizhevsky2012imagenet} due to its efficiency and generalization ability, in comparison to fully connected models \cite{ackley1985learning}. The success of convolution mainly comes from two properties: translation equivariance, and locality. Translation equivariance, although not exact \cite{zhang2019making}, aligns well with the nature of imaging and thus generalizes the model to different positions or to images of different sizes. Locality, on the other hand, reduces parameter counts and M-Adds. However, it makes modeling long range relations challenging.

A rich set of literature has discussed approaches to modeling long range interactions in convolutional neural networks (CNNs). Some employ atrous convolutions \cite{holschneider1990real,shensa1992discrete,papandreou2014untangling,deeplabv12015}, larger kernel \cite{peng2017large}, or image pyramids \cite{zhao2017pyramid,wang2019elastic}, either designed by hand or searched by algorithms \cite{zoph2017neural,chen2018searching,liu2019auto}. Another line of works adopts attention mechanisms. Attention shows its ability of modeling long range interactions in language modeling \cite{vaswani2017attention,wu2016google}, speech recognition \cite{chorowski2015attention,chan2016listen}, and neural captioning \cite{xu2015show}. Attention has since been extended to vision, giving significant boosts to image classification \cite{bello2019attention}, object detection \cite{hu2018relation}, semantic segmentation~\cite{huang2019ccnet}, video classification \cite{wang2018non}, and adversarial defense \cite{xie2019feature}. These works enrich CNNs with non-local or long-range attention modules.

Recently, stacking attention layers as stand-alone models without any spatial convolution has been proposed~\cite{parmar2019stand,hu2019local} and shown promising results. However, naive attention is computationally expensive, especially on large inputs. Applying local constraints to attention, proposed by \cite{parmar2019stand,hu2019local}, reduces the cost and enables building fully attentional models. However, local constraints limit model receptive field, which is crucial to tasks such as segmentation, especially on high-resolution inputs. In this work, we propose to adopt axial-attention~\cite{ho2019axial,huang2019ccnet}, which not only allows efficient computation, but recovers the large receptive field in stand-alone attention models. The core idea is to factorize 2D attention into two 1D attentions along height- and width-axis sequentially. Its efficiency enables us to attend over large regions and build models to learn long range or even global interactions. Additionally, most previous attention modules do not utilize positional information, which degrades attention's ability in modeling position-dependent interactions, like shapes or objects at multiple scales. Recent works~\cite{parmar2019stand,hu2019local,bello2019attention} introduce positional terms to attention, but in a context-agnostic way. In this paper, we augment the positional terms to be context-dependent, making our attention position-sensitive, with marginal costs.

We show the effectiveness of our axial-attention models on ImageNet~\cite{russakovsky2015imagenet} for classification, and on three datasets (COCO \cite{lin2014microsoft}, Mapillary Vistas \cite{neuhold2017mapillary}, and Cityscapes \cite{Cordts2016Cityscapes}) for panoptic segmentation \cite{kirillov2018panoptic}, instance segmentation, and semantic segmentation. In particular, on ImageNet, we build an Axial-ResNet by replacing the $3\times3$ convolution in all residual blocks \cite{he2016deep} with our position-sensitive axial-attention layer, and we further make it fully attentional \cite{parmar2019stand} by adopting axial-attention layers in the `stem'. As a result, our Axial-ResNet attains state-of-the-art results among stand-alone attention models on ImageNet. For segmentation tasks, we convert Axial-ResNet to Axial-DeepLab by replacing the backbones in Panoptic-DeepLab \cite{cheng2019panopticworkshop}. On COCO \cite{lin2014microsoft}, our Axial-DeepLab outperforms the current {\it bottom-up} state-of-the-art, Panoptic-DeepLab~\cite{cheng2019panoptic}, by 2.8\% PQ on test-dev set. We also show state-of-the-art segmentation results on Mapillary Vistas \cite{neuhold2017mapillary}, and Cityscapes \cite{Cordts2016Cityscapes}.

To summarize, our contributions are four-fold:
\begin{itemize}
  \item The proposed method is the first attempt to build stand-alone attention models with large or global receptive field.
  \item We propose position-sensitive attention layer that makes better use of positional information without adding much computational cost.
  \item We show that axial attention works well, not only as a stand-alone model on image classification, but also as a backbone on panoptic segmentation, instance segmentation, and segmantic segmentation.
  \item Our Axial-DeepLab improves significantly over bottom-up state-of-the-art on COCO, achieving comparable performance of two-stage methods. We also surpass previous state-of-the-art methods on Mapillary Vistas and Cityscapes.
\end{itemize}

\section{Related Work}
\label{sec:relatedwork}
{\bf Top-down panoptic segmentation:} Most state-of-the-art panoptic segmentation models employ a two-stage approach where object proposals are firstly generated followed by sequential processing of each proposal. We refer to such approaches as top-down or proposal-based methods. Mask R-CNN \cite{he2017mask} is commonly deployed in the pipeline for instance segmentation, paired with  a light-weight stuff segmentation branch. For example, Panoptic FPN \cite{kirillov2019panoptic} incorporates a semantic segmentation head to Mask R-CNN \cite{he2017mask}, while Porzi \etal \cite{porzi2019seamless} append a light-weight DeepLab-inspired module \cite{chen2018deeplabv2} to the multi-scale features from FPN \cite{lin2017feature}. Additionally, some extra modules are designed to resolve the overlapping instance predictions by Mask R-CNN. TASCNet \cite{li2018learning} and AUNet \cite{li2018attention} propose a module to guide the fusion between `thing' and `stuff' predictions, while Liu \etal \cite{liu2019e2e} adopt a Spatial Ranking module.  UPSNet \cite{xiong2019upsnet} develops an efficient parameter-free panoptic head for fusing `thing' and `stuff', which is further explored by Li~\etal \cite{li2020unifying} for end-to-end training of panoptic segmentation models. AdaptIS \cite{sofiiuk2019adaptis} uses point proposals to generate instance masks.

{\bf Bottom-up panoptic segmentation:} In contrast to top-down approaches, bottom-up or proposal-free methods for panoptic segmentation typically start with the semantic segmentation prediction followed by grouping `thing' pixels into clusters to obtain instance segmentation. DeeperLab \cite{yang2019deeperlab} predicts bounding box four corners and object centers for class-agnostic instance segmentation. SSAP \cite{gao2019ssap} exploits the pixel-pair affinity pyramid \cite{liu2018affinity} enabled by an efficient graph partition method \cite{keuper2015efficient}. BBFNet \cite{bonde2020towards} obtains instance segmentation results by Watershed transform \cite{vincent1991watersheds,bai2017deep} and Hough-voting \cite{ballard1981generalizing,leibe2004combined}. Recently, Panoptic-DeepLab \cite{cheng2019panoptic}, a simple, fast, and strong approach for bottom-up panoptic segmentation, employs a class-agnostic instance segmentation branch involving a simple instance center regression \cite{kendall2018multi,uhrig2018box2pix,neven2019instance}, coupled with DeepLab semantic segmentation outputs  \cite{deeplabv12015,chen2017deeplabv3,deeplabv3plus2018}. Panoptic-DeepLab has achieved state-of-the-art results on several benchmarks, and our method builds on top of it.

{\bf Self-attention:} Attention, introduced by~\cite{bahdanau2014neural} for the encoder-decoder in a neural sequence-to-sequence model, is developed to capture correspondence of tokens between two sequences. In contrast, self-attention is defined as applying attention to a single context instead of across multiple modalities. Its ability to directly encode long-range interactions and its parallelizability, has led to state-of-the-art performance for various tasks~\cite{vaswani2017attention,huang2019music,devlin2018bert,parmar2018image,shaw2018self,dai2019transformer,li2020neural}. Recently, self-attention has been applied to computer vision, by augmenting CNNs with non-local or long-range modules. 
Non-local neural networks~\cite{wang2018non} show that self-attention is an instantiation of non-local means~\cite{buades2005non} and achieve gains on many vision tasks such as video classification and object detection. Additionally, \cite{chen20182,bello2019attention} show improvements on image classification by combining features from self-attention and convolution. State-of-the-art results on video action recognition tasks~\cite{chen20182} are also achieved in this way. On semantic segmentation, self-attention is developed as a context aggregation module that captures multi-scale context~\cite{huang2019ccnet,fu2019dual,zhu2019asymmetric,zhu2019empirical}. Efficient attention methods are proposed to reduce its complexity~\cite{shen2018efficient,huang2019ccnet,li2020neural}. Additionally, CNNs augmented with non-local means \cite{buades2005non} are shown to be more robust to adversarial attacks~\cite{xie2019feature}. Besides discriminative tasks, self-attention is also applied to generative modeling of images~\cite{zhang2018self,brock2019large,ho2019axial}.
Recently, \cite{parmar2019stand,hu2019local} show that self-attention layers alone could be stacked to form a fully attentional model by restricting the receptive field of self-attention to a {\it local} square region. Encouraging results are shown on both image classification and object detection. In this work, we follow this direction of research and propose a stand-alone self-attention model with large or global receptive field, making self-attention models {\it non-local} again. Our models are evaluated on bottom-up panoptic segmentation and show significant improvements.
\section{Method}
\label{sec:model}
We begin by formally introducing our position-sensitive self-attention mechanism. Then, we discuss how it is applied to axial-attention and how we build stand-alone Axial-ResNet and Axial-DeepLab with axial-attention layers.

\subsection{Position-Sensitive Self-Attention}
{\bf Self-Attention:} Self-attention mechanism is usually applied to vision models as an add-on to augment CNNs outputs  \cite{wang2018non,zhang2018self,huang2019ccnet}. Given an input feature map $x \in \mathbb{R}^{h \times w \times d_{in}}$ with height $h$, width $w$, and channels $d_{in}$, the output at position $o=(i,j)$, $y_{o} \in \mathbb{R}^{d_{out}}$, is computed by pooling over the projected input as:
\begin{equation}
    y_{o} = \sum_{p \in \mathcal{N}} \text{softmax}_{p}(q_{o}^T k_{p})v_{p}
\end{equation}
where $\mathcal{N}$ is the whole location lattice, and queries $q_{o}=W_Q x_{o}$, keys $k_{o}=W_K x_{o}$, values $v_{o}=W_V x_{o}$ are all linear projections of the input $x_{o}~\forall o \in \mathcal{N}$. $W_Q, W_K \in \mathbb{R}^{d_{q} \times d_{in}}$ and $W_V \in \mathbb{R}^{d_{out} \times d_{in}}$ are all learnable matrices. The $\text{softmax}_{p}$ denotes a softmax function applied to all possible $p=(a, b)$ positions, which in this case is also the whole 2D lattice.

This mechanism pools values $v_{p}$ globally based on affinities $x_{o}^T W_Q^T W_K x_{p}$, allowing us to capture related but non-local context in the whole feature map, as opposed to convolution which only captures local relations.

However, self-attention is extremely expensive to compute ($\mathcal{O}(h^2 w^2)$) when the spatial dimension of the input is large, restricting its use to only high levels of a CNN (\ie, downsampled feature maps) or small images. Another drawback is that the global pooling does not exploit positional information, which is critical to capture spatial structures or shapes in vision tasks.

These two issues are mitigated in \cite{parmar2019stand} by adding local constraints and positional encodings to self-attention. For each location $o$, a local $m\times m$ square region is extracted to serve as a memory bank for computing the output $y_{o}$. This significantly reduces its computation to $\mathcal{O}(h w m^2)$, allowing self-attention modules to be deployed as stand-alone layers to form a fully self-attentional neural network. Additionally, a learned relative positional encoding term is incorporated into the affinities, yielding a dynamic prior of where to look at in the receptive field (\ie, the local $m\times m$ square region). Formally, \cite{parmar2019stand} proposes
\begin{equation}
\label{eqn:standalone}
    y_{o} = \sum_{p \in \mathcal{N}_{m \times m}(o)} \text{softmax}_{p}(q_{o}^T k_{p} + q_{o}^T r_{p-o}) v_{p}
\end{equation}
where $\mathcal{N}_{m \times m}(o)$ is the local $m \times m$ square region centered around location $o=(i,j)$, and the learnable vector $r_{p-o} \in \mathbb{R}^{d_{q}}$ is the added relative positional encoding. The inner product $q_{o}^T r_{p-o}$ measures the compatibility from location $p=(a,b)$ to location $o=(i,j)$. We do not consider absolute positional encoding $q_{o}^T r_{p}$, because they do not generalize well compared to the relative counterpart~\cite{parmar2019stand}. In the following paragraphs, we drop the term relative for conciseness.

In practice, $d_q$ and $d_{out}$ are much smaller than $d_{in}$, and one could extend single-head attention in \equref{eqn:standalone} to multi-head attention to capture a mixture of affinities. In particular, multi-head attention is computed by applying $N$ single-head attentions in parallel on $x_{o}$ (with different $W_Q^n, W_K^n, W_V^n, \forall n \in \{1, 2, \dots, N\}$ for the $n$-th head), and then obtaining the final output $z_{o}$ by concatenating the results from each head, \ie, $z_{o}=\text{concat}_n(y^n_{o})$. Note that positional encodings are often shared across heads, so that they introduce marginal extra parameters.

{\bf Position-Sensitivity:} We notice that previous positional bias only depends on the query pixel $x_{o}$, not the key pixel $x_{p}$. However, the keys $x_{p}$ could also have information about which location to attend to. We therefore add a key-dependent positional bias term $k_{p}^T r^{k}_{p-o}$, besides the query-dependent bias $q_{o}^T r^{q}_{p-o}$.

Similarly, the values $v_{p}$ do not contain any positional information in \equref{eqn:standalone}. In the case of large receptive fields or memory banks, it is unlikely that $y_{o}$ contains the precise location from which $v_{p}$ comes. Thus, previous models have to trade-off between using smaller receptive fields (\ie, small $m\times m$ regions) and throwing away precise spatial structures. In this work, we enable the output $y_{o}$ to retrieve relative positions $r^{v}_{p-o}$, besides the content $v_{p}$, based on query-key affinities $q_{o}^T k_{p}$. Formally,
\begin{equation}
    y_{o} = \sum_{p \in \mathcal{N}_{m \times m}(o)} \text{softmax}_{p}(q_{o}^T k_{p} + q_{o}^T r^{q}_{p-o} + k_{p}^T r^{k}_{p-o}) (v_{p} + r^{v}_{p-o})
\end{equation}
where the learnable $r^{k}_{p-o} \in \mathbb{R}^{d_{q}}$ is the positional encoding for keys, and $r^{v}_{p-o} \in \mathbb{R}^{d_{out}}$ is for values. Both vectors do not introduce many parameters, since they are shared across attention heads in a layer, and the number of local pixels $|\mathcal{N}_{m \times m}(o)|$ is usually small.

We call this design {\it position-sensitive} self-attention, which captures long range interactions with precise positional information at a reasonable computation overhead, as verified in our experiments.

\subsection{Axial-Attention}

The local constraint, proposed by the stand-alone self-attention models \cite{parmar2019stand}, significantly reduces the computational costs in vision tasks and enables building fully self-attentional model. However, such constraint sacrifices the global connection, making attention's receptive field no larger than a depthwise convolution with the same kernel size. Additionally, the local self-attention, performed in local square regions, still has complexity quadratic to the region length, introducing another hyper-parameter to trade-off between performance and computation complexity. In this work, we propose to adopt axial-attention \cite{huang2019ccnet,ho2019axial} in stand-alone self-attention, ensuring both global connection and efficient computation. Specifically, we first define an axial-attention layer on the {\it width}-axis of an image as simply a one dimensional {\it position-sensitive} self-attention, and use the similar definition for the {\it height}-axis. To be concrete, the axial-attention layer along the width-axis is defined as follows.

\begin{figure}[!t]
    \centering
    \includegraphics[height=4.9cm]{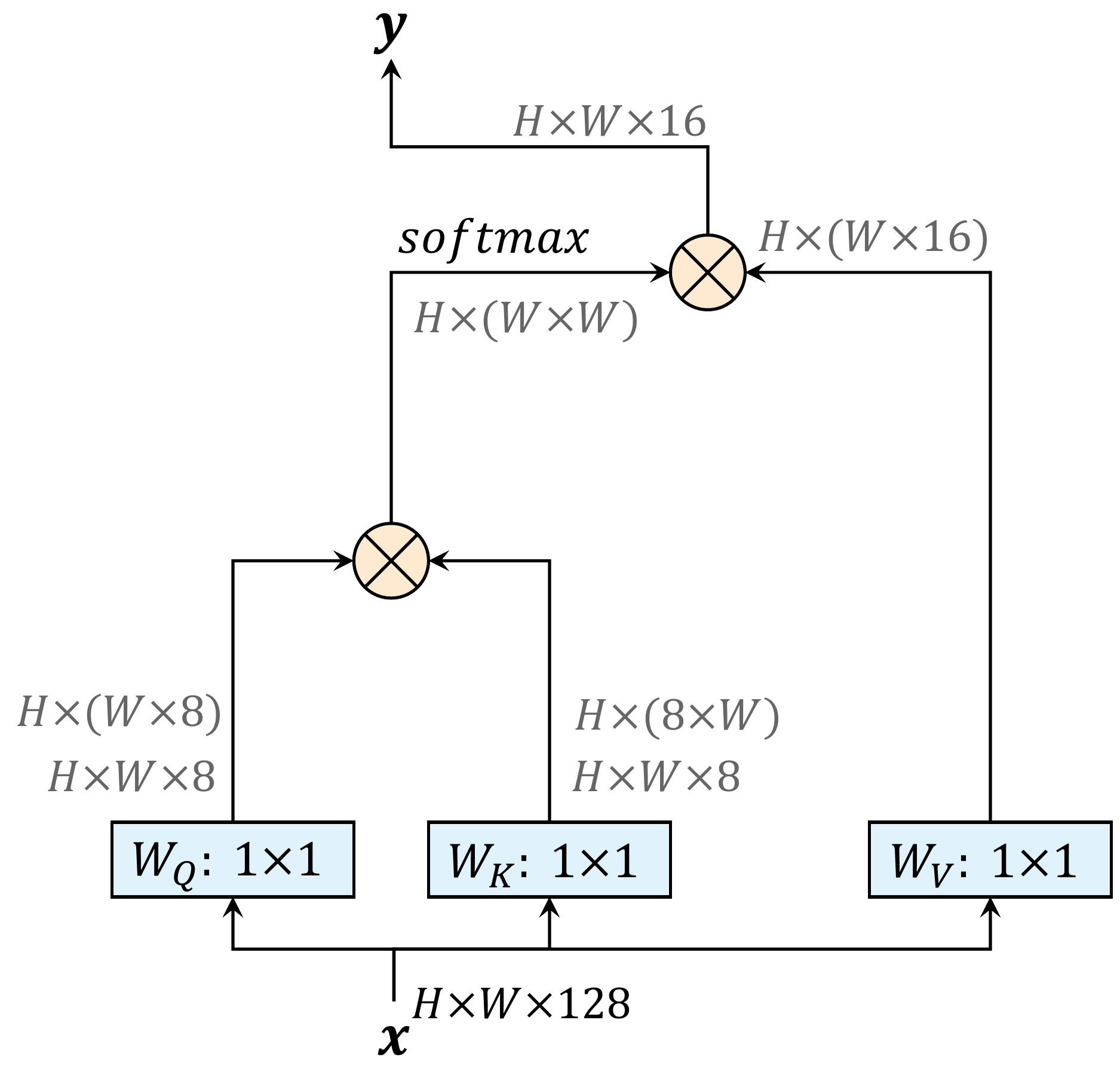}
    ~
    \includegraphics[height=4.9cm]{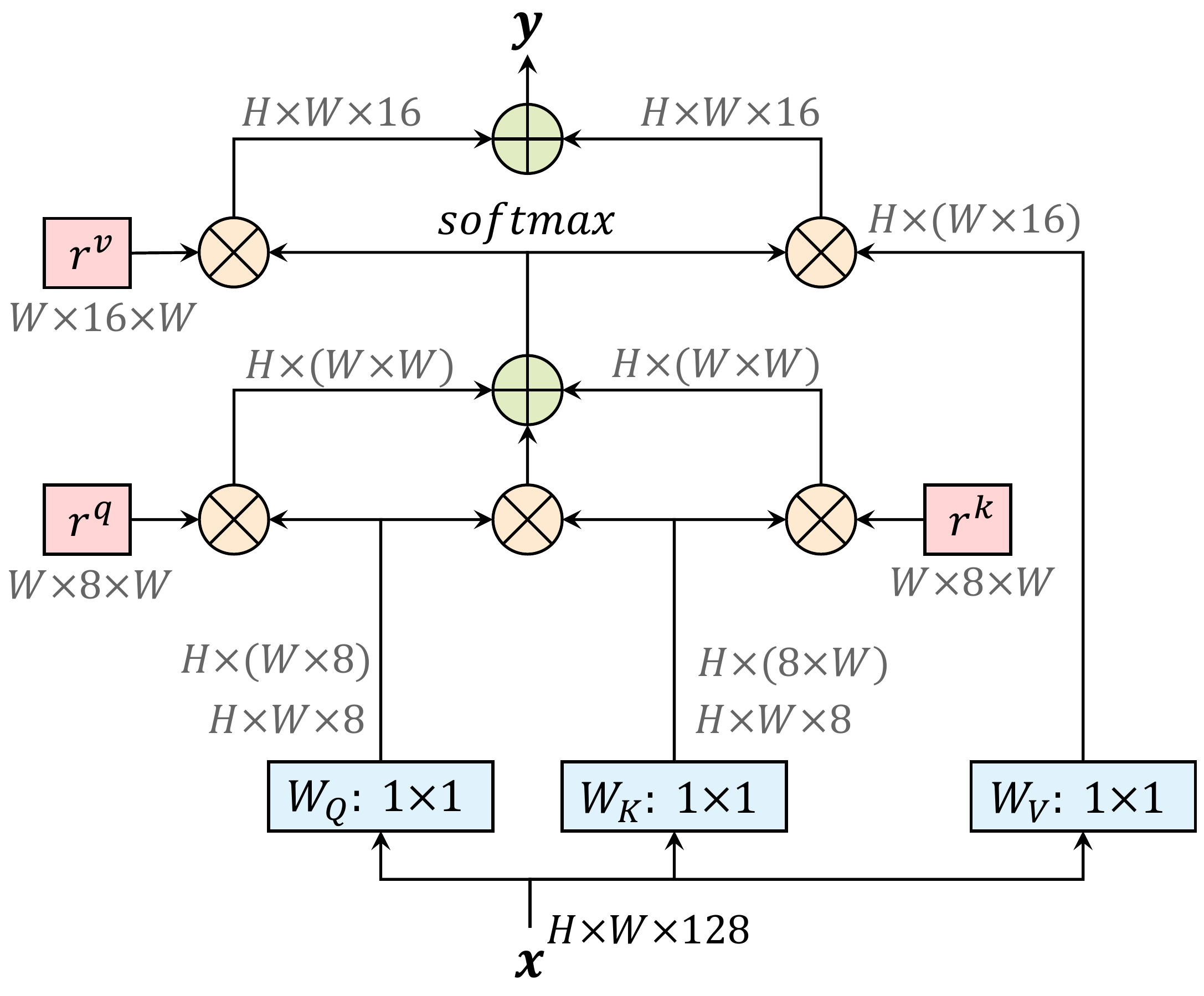}
    \caption{A non-local block (left) \vs our position-sensitive axial-attention applied along the width-axis (right). ``$\otimes$'' denotes matrix multiplication, and ``$\oplus$'' denotes element-wise sum. The softmax is performed on the last axis. Blue boxes denote $1\times1$ convolutions, and red boxes denote relative positional encoding. The channels $d_{in}=128$, $d_{q}=8$, and $d_{out}=16$ is what we use in the first stage of ResNet after `stem'}
    \label{fig:axial_layer}
\end{figure}
\begin{equation}
    y_{o} = \sum_{p \in \mathcal{N}_{1 \times m}(o)} \text{softmax}_{p}(q_{o}^T k_{p} + q_{o}^T r^{q}_{p-o} + k_{p}^T r^{k}_{p-o}) (v_{p} + r^{v}_{p-o})
\end{equation}
One axial-attention layer propagates information along one particular axis. To capture global information, we employ two axial-attention layers consecutively for the height-axis and width-axis, respectively. Both of the axial-attention layers adopt the multi-head attention mechanism, as described above.

Axial-attention reduces the complexity to $\mathcal{O}(hwm)$. This enables global receptive field, which is achieved by setting the span $m$ directly to the whole input features. Optionally, one could also use a fixed $m$ value, in order to reduce memory footprint on huge feature maps.
\begin{figure}[!t]
    \centering
    \includegraphics[width=0.9\textwidth]{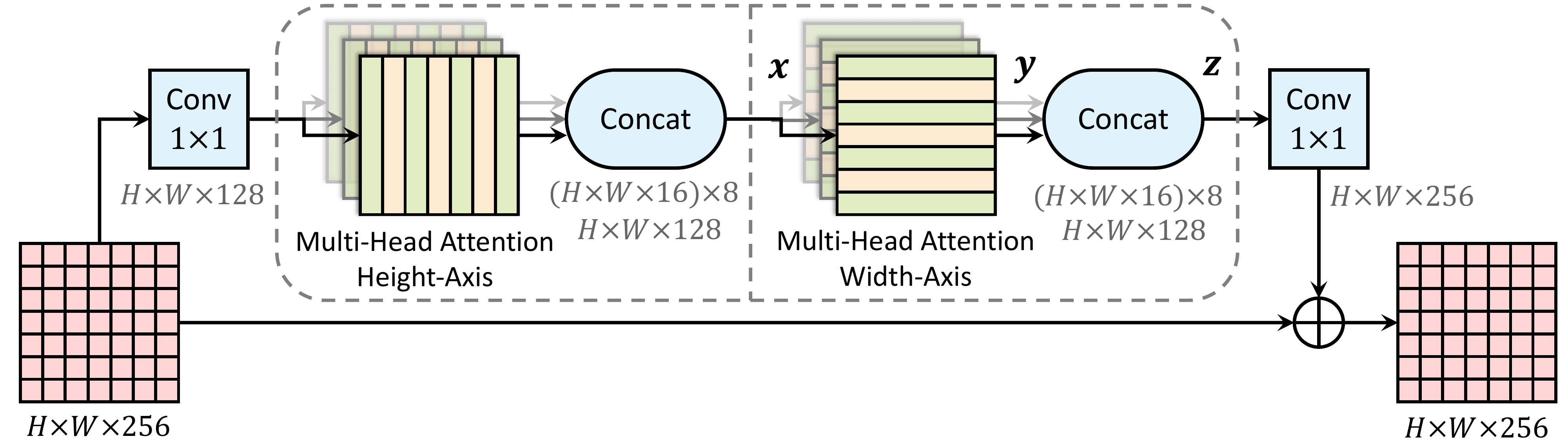}
    \caption{An axial-attention block, which consists of two axial-attention layers operating along height- and width-axis sequentially. The channels $d_{in}=128$, $d_{out}=16$ is what we use in the first stage of ResNet after `stem'. We employ $N=8$ attention heads}
    \label{fig:axial_block}
\end{figure}

{\bf Axial-ResNet:} To transform a ResNet \cite{he2016deep} to an {\it Axial}-ResNet, we replace the $3\times3$ convolution in the residual bottleneck block by two multi-head axial-attention layers (one for height-axis and the other for width-axis). Optional striding is performed on each axis after the corresponding axial-attention layer. The two $1\times1$ convolutions are kept to shuffle the features. This forms our (residual) axial-attention block, as illustrated in \figref{fig:axial_block}, which is stacked multiple times to obtain Axial-ResNets. Note that we do not use a $1\times1$ convolution in-between the two axial-attention layers, since matrix multiplications ($W_Q, W_K, W_V$) follow immediately. Additionally, the stem (\ie, the first strided $7\times7$ convolution and $3\times3$ max-pooling) in the original ResNet is kept, resulting in a {\it conv-stem} model where convolution is used in the first layer and attention layers are used everywhere else. In {\it conv-stem} models, we set the span $m$ to the whole input from the first block, where the feature map is 56$\times$56.

In our experiments, we also build a full axial-attention model, called Full Axial-ResNet, which further applies axial-attention to the stem. Instead of designing a special spatially-varying attention stem \cite{parmar2019stand}, we simply stack three axial-attention bottleneck blocks. In addition, we adopt local constraints (\ie, a local $m\times m$ square region as in \cite{parmar2019stand}) in the first few blocks of Full Axial-ResNets, in order to reduce computational cost.

{\bf Axial-DeepLab:} To further convert Axial-ResNet to Axial-DeepLab for segmentation tasks, we make several changes as discussed below.

Firstly, to extract dense feature maps, DeepLab \cite{deeplabv12015} changes the stride and atrous rates of the last one or two stages in ResNet \cite{he2016deep}. Similarly, we remove the stride of the last stage but we do not implement the `atrous' attention module, since our axial-attention already captures global information for the whole input. In this work, we extract feature maps with output stride (\ie, the ratio of input resolution to the final backbone feature resolution) 16. We do not pursue output stride 8, since it is computationally expensive.

Secondly, we do not adopt the atrous spatial pyramid pooling module (ASPP) \cite{chen2018deeplabv2,chen2017deeplabv3}, since our axial-attention block could also efficiently encode the multi-scale or global information. We show in the experiments that our Axial-DeepLab without ASPP outperforms Panoptic-DeepLab \cite{cheng2019panoptic} with and without ASPP.

Lastly, following Panoptic-DeepLab~\cite{cheng2019panoptic}, we adopt exactly the same stem~\cite{szegedy2016rethinking} of three convolutions, dual decoders, and prediction heads. The heads produce semantic segmentation and class-agnostic instance segmentation, and they are merged by majority voting \cite{yang2019deeperlab} to form the final panoptic segmentation.

In cases where the inputs are extremely large (\eg, $2177 \times 2177$) and memory is constrained, we resort to a large span $m=65$ in all our axial-attention blocks. Note that we do not consider the axial span as a hyper-parameter because it is already sufficient to cover long range or even global context on several datasets, and setting a smaller span does not significantly reduce M-Adds.

\section{Experimental Results}
\label{sec:experiments}
We conduct experiments on four large-scale datasets. We first report results with our Axial-ResNet on ImageNet \cite{russakovsky2015imagenet}. We then convert the ImageNet pretrained Axial-ResNet to Axial-DeepLab, and report results on COCO \cite{lin2014microsoft}, Mapillary Vistas \cite{neuhold2017mapillary}, and Cityscapes \cite{Cordts2016Cityscapes} for panoptic segmentation, evaluated by panoptic quality (PQ) \cite{kirillov2018panoptic}. We also report average precision (AP) for instance segmentation, and mean IoU for semantic segmentation on Mapillary Vistas and Cityscapes. Our models are trained using TensorFlow \cite{tensorflow-osdi2016} on 128 TPU cores for ImageNet and 32 cores for panoptic segmentation.

{\bf Training protocol:} On ImageNet, we adopt the same training protocol as \cite{parmar2019stand} for a fair comparison, except that we use batch size 512 for Full Axial-ResNets and 1024 for all other models, with learning rates scaled accordingly~\cite{goyal2017accurate}.

For panoptic segmentation, we strictly follow Panoptic-DeepLab~\cite{cheng2019panoptic}, except using a linear warm up Radam \cite{liu2019variance} Lookahead \cite{zhang2019lookahead} optimizer (with the same learning rate 0.001). All our results on panoptic segmentation use this setting. We note this change does not improve the results, but smooths our training curves. Panoptic-DeepLab yields similar result in this setting.

\subsection{ImageNet}

For ImageNet, we build Axial-ResNet-L from ResNet-50 \cite{he2016deep}. In detail, we set $d_{in}=128$, $d_{out}=2d_{q}=16$ for the first stage after the `stem'. We double them when spatial resolution is reduced by a factor of 2~\cite{simonyan2014very}. Additionally, we multiply all the channels ~\cite{howard2017mobilenets,sandler2018mobilenetv2,howard2019searching} by 0.5, 0.75, and 2, resulting in Axial-ResNet-\{S, M, XL\}, respectively. Finally, {\it Stand-Alone} Axial-ResNets are further generated by replacing the `stem' with three axial-attention blocks where the first block has stride 2. Due to the computational cost introduced by the early layers, we set the axial span $m=15$ in all blocks of Stand-Alone Axial-ResNets. We always use $N=8$ heads~\cite{parmar2019stand}. In order to avoid careful initialization of $W_Q, W_K, W_V, r^q, r^k, r^v$, we use batch normalizations~\cite{ioffe2015batch} in all attention layers.

\tabref{tab:imagenet} summarizes our ImageNet results. The baselines ResNet-50 \cite{he2016deep} (done by \cite{parmar2019stand}) and Conv-Stem + Attention \cite{parmar2019stand} are also listed. In the conv-stem setting, adding BN to attention layers of \cite{parmar2019stand} slightly improves the performance by 0.3\%. Our proposed position-sensitive self-attention (Conv-Stem + PS-Attention) further improves the performance by 0.4\% at the cost of extra marginal computation. Our Conv-Stem + Axial-Attention performs on par with Conv-Stem + Attention \cite{parmar2019stand} while being more parameter- and computation-efficient. When comparing with other full self-attention models, our Full Axial-Attention outperforms Full Attention~\cite{parmar2019stand} by 0.5\%, while being 1.44$\times$ more parameter-efficient and 1.09$\times$ more computation-efficient.

\begin{figure}[!t]
    \centering
    \includegraphics[width=0.47\textwidth]{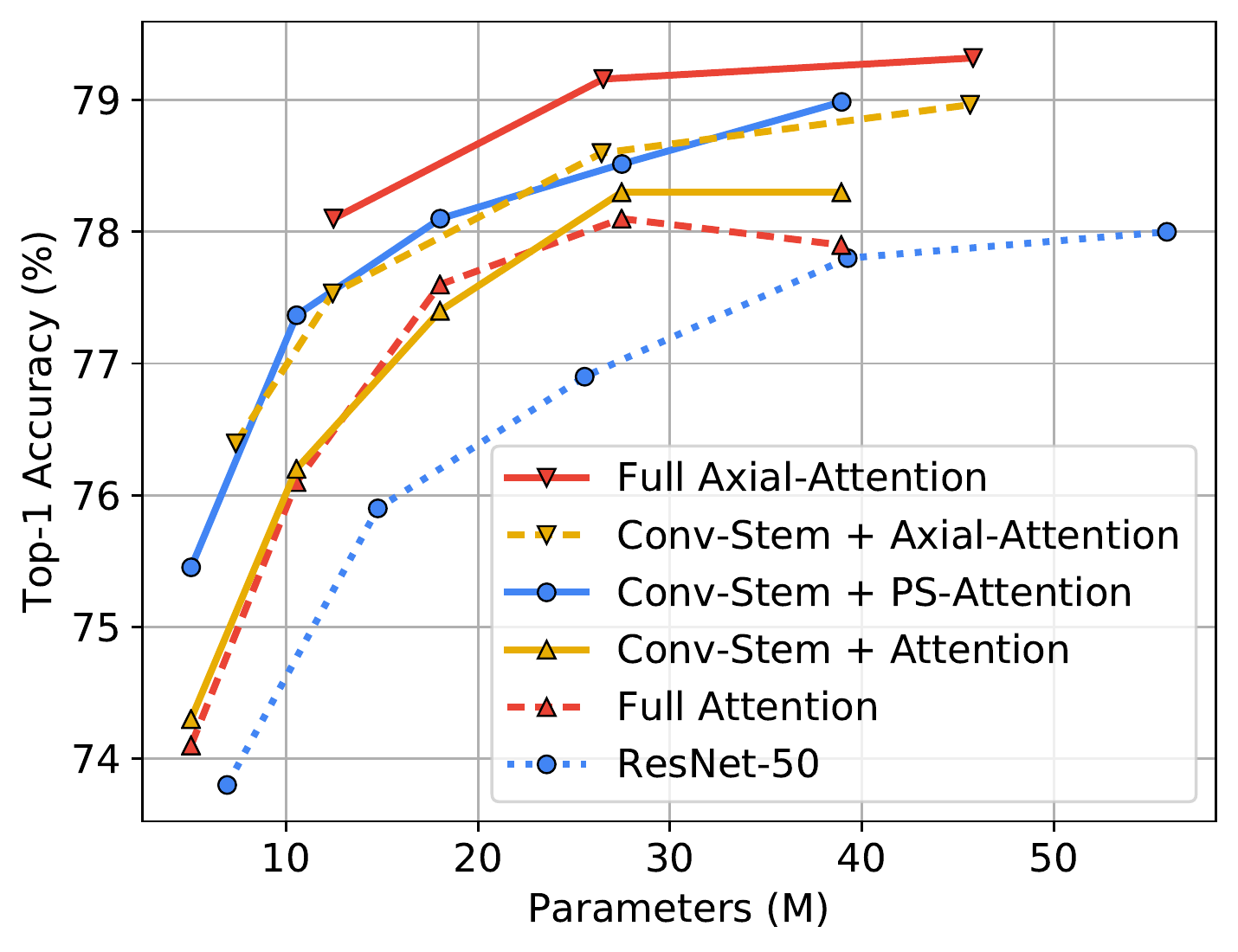}
    ~
    \includegraphics[width=0.47\textwidth]{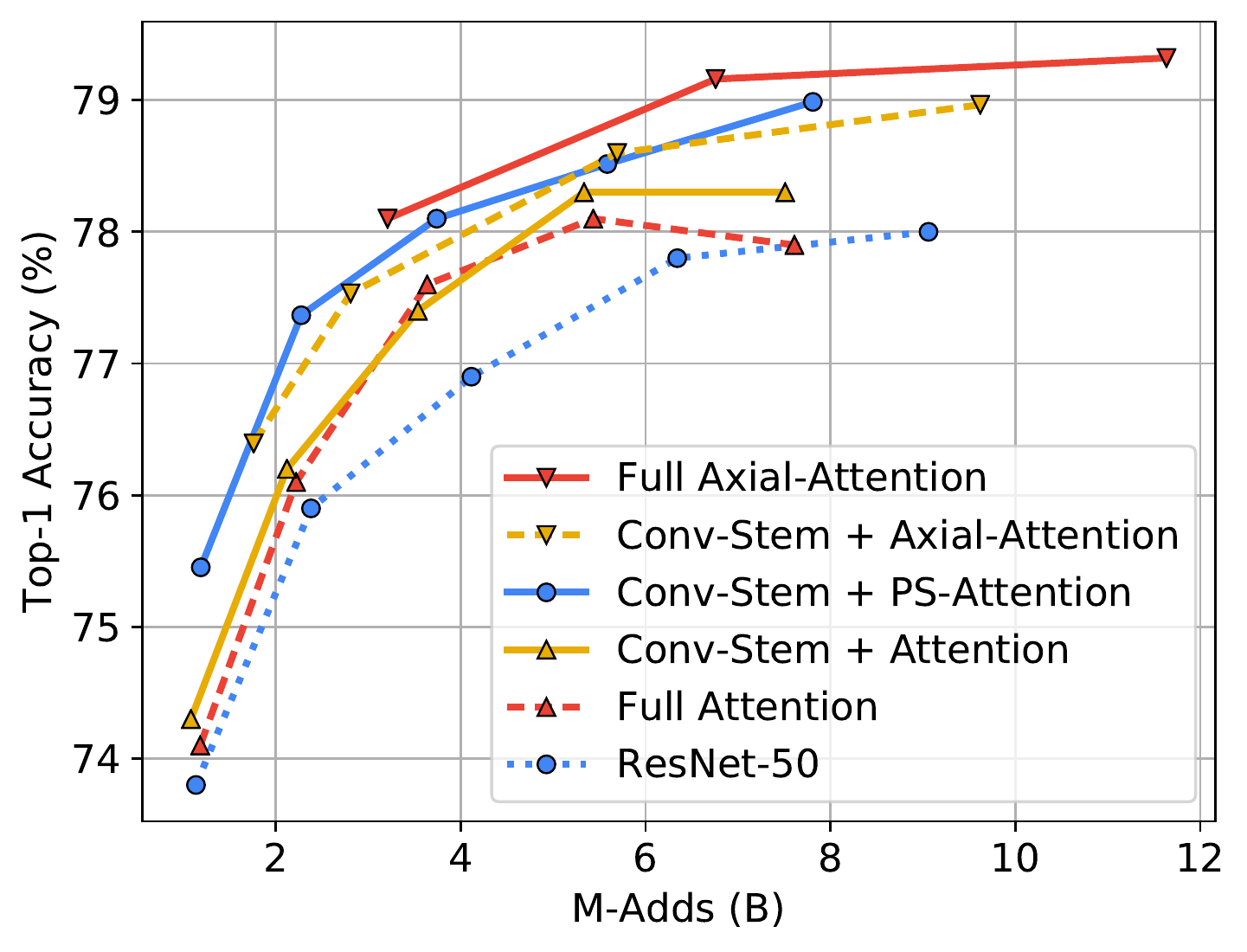}
    \caption{Comparing parameters and M-Adds against accuracy on ImageNet classification. Our position-sensitive self-attention (Conv-Stem + PS-Attention) and axial-attention (Conv-Stem + Axial-Attention) consistently outperform ResNet-50~\cite{he2016deep,parmar2019stand} and attention models~\cite{parmar2019stand} (both Conv-Stem + Attention, and Full Attention), across a range of network widths (\ie, different channels). Our Full Axial-Attention works the best in terms of both parameters and M-Adds}
    \label{fig:params_flops}
\end{figure}

Following \cite{parmar2019stand}, we experiment with different network widths (\ie, Axial-ResNets-\{S,M,L,XL\}), exploring the trade-off between accuracy, model parameters, and computational cost (in terms of M-Adds). As shown in \figref{fig:params_flops}, our proposed Conv-Stem + PS-Attention and Conv-Stem + Axial-Attention already outperforms ResNet-50~\cite{he2016deep,parmar2019stand} and attention models~\cite{parmar2019stand} (both Conv-Stem + Attention, and Full Attention) at all settings. Our Full Axial-Attention further attains the best accuracy-parameter and accuracy-complexity trade-offs.

\begin{table}[!t]
\setlength{\tabcolsep}{0.72em}
    \centering
    \caption{ImageNet validation set results. \textbf{BN:} Use batch normalizations in attention layers. \textbf{PS:} Our position-sensitive self-attention. \textbf{Full:} Stand-alone self-attention models without spatial convolutions}
    \label{tab:imagenet}
    \scalebox{1.0}{
    \begin{tabular}{l|c|c|c|c|c|c}
        \toprule[0.2em]
        Method & BN & PS & Full & Params & M-Adds & Top-1 \\
        \toprule[0.2em]
        \multicolumn{7}{c}{Conv-Stem methods} \\
        \midrule
        ResNet-50~\cite{he2016deep,parmar2019stand} & & & & 25.6M & 4.1B & 76.9 \\
        Conv-Stem + Attention~\cite{parmar2019stand} & & & & 18.0M & 3.5B & 77.4 \\
        \midrule
        Conv-Stem + Attention & \ding{51} & & & 18.0M & 3.5B & 77.7 \\
        Conv-Stem + PS-Attention & \ding{51} & \ding{51} & & 18.0M & 3.7B & 78.1 \\
        Conv-Stem + Axial-Attention & \ding{51} & \ding{51} & & 12.4M & 2.8B & 77.5 \\
        \midrule
        \multicolumn{7}{c}{Fully self-attentional methods} \\
        \midrule 
        LR-Net-50~\cite{hu2019local} & & & \ding{51} & 23.3M & 4.3B & 77.3 \\
        Full Attention~\cite{parmar2019stand} & & &\ding{51} & 18.0M & 3.6B & 77.6 \\
        Full Axial-Attention & \ding{51} & \ding{51} & \ding{51} & \textbf{12.5M} & \textbf{3.3B} & \textbf{78.1} \\
        \bottomrule[0.1em]
    \end{tabular}
    }
\end{table}

\subsection{COCO}

The ImageNet pretrained Axial-ResNet model variants (with different channels) are then converted to Axial-DeepLab model variant for panoptic segmentation tasks. We first demonstrate the effectiveness of our Axial-DeepLab on the challenging COCO dataset \cite{lin2014microsoft}, which contains objects with various scales (from less than $32\times32$ to larger than $96\times96$).

{\bf Val set:} In \tabref{tab:coco_val}, we report our validation set results and compare with other bottom-up panoptic segmentation methods, since our method also belongs to the bottom-up family. As shown in the table, our {\it single-scale} Axial-DeepLab-S outperforms DeeperLab \cite{yang2019deeperlab} by 8\% PQ, {\it multi-scale} SSAP \cite{gao2019ssap} by 5.3\% PQ, and {\it single-scale} Panoptic-DeepLab by 2.1\% PQ. Interestingly, our {\it single-scale} Axial-DeepLab-S also outperforms {\it multi-scale} Panoptic-DeepLab by 0.6\% PQ while being {\bf 3.8$\times$} parameter-efficient and {\bf 27$\times$} computation-efficient (in M-Adds). Increasing the backbone capacity (via large channels) continuously improves the performance. Specifically, our {\it multi-scale} Axial-DeepLab-L attains 43.9\% PQ, outperforming Panoptic-DeepLab \cite{cheng2019panoptic} by 2.7\% PQ.

    \begin{table}[!t]
    \caption{COCO val set. {\bf MS:} Multi-scale inputs}
    \label{tab:coco_val}
    \setlength{\tabcolsep}{0.31em}
        \centering
        \scalebox{1.0}{
        \begin{tabular}{l|c|c|cc|ccc}
            \toprule[0.2em]
            Method & Backbone & MS & Params & M-Adds & PQ & PQ\textsuperscript{Th} & PQ\textsuperscript{St} \\
            \toprule[0.2em]
            DeeperLab~\cite{yang2019deeperlab} & Xception-71 & & & & 33.8 & - & - \\
            SSAP~\cite{gao2019ssap} & ResNet-101 & \ding{51} & & & 36.5 & - & - \\
            Panoptic-DeepLab~\cite{cheng2019panoptic} & Xception-71 & & 46.7M & 274.0B & 39.7 & 43.9 & 33.2 \\
            Panoptic-DeepLab~\cite{cheng2019panoptic} & Xception-71 & \ding{51} & 46.7M & 3081.4B & 41.2 & 44.9 & 35.7 \\
            \midrule \midrule
            Axial-DeepLab-S & Axial-ResNet-S & & 12.1M & 110.4B & 41.8 & 46.1 & 35.2 \\
            Axial-DeepLab-M & Axial-ResNet-M & & 25.9M & 209.9B & 42.9 & 47.6 & 35.8 \\
            Axial-DeepLab-L & Axial-ResNet-L & & 44.9M & 343.9B & 43.4 & 48.5 & 35.6 \\
            Axial-DeepLab-L & Axial-ResNet-L & \ding{51} & 44.9M & 3867.7B & 43.9 & 48.6 & 36.8 \\
            \bottomrule[0.1em]
        \end{tabular}
        }
    \end{table}

{\bf Test-dev set:} As shown in \tabref{tab:coco_test}, our Axial-DeepLab variants show consistent improvements with larger backbones. Our {\it multi-scale} Axial-DeepLab-L attains the performance of 44.2\% PQ, outperforming DeeperLab~\cite{yang2019deeperlab} by 9.9\% PQ, SSAP~\cite{gao2019ssap} by 7.3\% PQ, and Panoptic-DeepLab~\cite{cheng2019panoptic} by 2.8\% PQ, setting a new state-of-the-art among bottom-up approaches. We also list several top-performing methods adopting the top-down approaches in the table for reference.

    \begin{table}[!t]
    \caption{COCO test-dev set. {\bf MS:} Multi-scale inputs}
    \label{tab:coco_test}
    \setlength{\tabcolsep}{0.9em}
        \centering
        \scalebox{1.0}{
        \begin{tabular}{l|c|c|ccc}
            \toprule[0.2em]
            Method & Backbone & MS & PQ & PQ\textsuperscript{Th} & PQ\textsuperscript{St} \\
            \toprule[0.2em]
            \multicolumn{5}{c}{Top-down panoptic segmentation methods} \\
            \midrule
            TASCNet~\cite{li2018learning} & ResNet-50 & & 40.7 & 47.0 & 31.0 \\
            Panoptic-FPN~\cite{kirillov2019panoptic} & ResNet-101 & & 40.9 & 48.3 & 29.7  \\
            AdaptIS~\cite{sofiiuk2019adaptis} & ResNeXt-101 & \cmark & 42.8 & 53.2 & 36.7 \\
            AUNet~\cite{li2018attention} & ResNeXt-152 & & 46.5 & 55.8 & 32.5 \\
            UPSNet~\cite{xiong2019upsnet} & DCN-101 \cite{dai2017deformable} & \cmark & 46.6 & 53.2 & 36.7 \\
            Li~\etal~\cite{li2020unifying} & DCN-101 \cite{dai2017deformable} & & 47.2 & 53.5 & 37.7 \\
            SpatialFlow~\cite{chen2019spatialflow} & DCN-101 \cite{dai2017deformable} & \cmark & 47.3 & 53.5 & 37.9 \\
            SOGNet~\cite{yang2019sognet} & DCN-101 \cite{dai2017deformable} & \cmark & 47.8 & - & - \\
            \midrule
            \multicolumn{5}{c}{Bottom-up panoptic segmentation methods} \\
            \midrule
            DeeperLab~\cite{yang2019deeperlab} & Xception-71 & & 34.3 & 37.5 & 29.6 \\
            SSAP~\cite{gao2019ssap} & ResNet-101 & \cmark & 36.9 & 40.1 & 32.0 \\
            Panoptic-DeepLab~\cite{cheng2019panoptic} & Xception-71 & \cmark & 41.4 & 45.1 & 35.9 \\
            \midrule \midrule
            Axial-DeepLab-S & Axial-ResNet-S & & 42.2 & 46.5 & 35.7 \\
            Axial-DeepLab-M & Axial-ResNet-M & & 43.2 & 48.1 & 35.9 \\
            Axial-DeepLab-L & Axial-ResNet-L & & 43.6 & 48.9 & 35.6 \\
            Axial-DeepLab-L & Axial-ResNet-L & \cmark & 44.2 & 49.2 & 36.8 \\
            \bottomrule[0.1em]
        \end{tabular}
        }
    \end{table}

\begin{figure}[!t]
    \centering
    \includegraphics[width=0.43\textwidth]{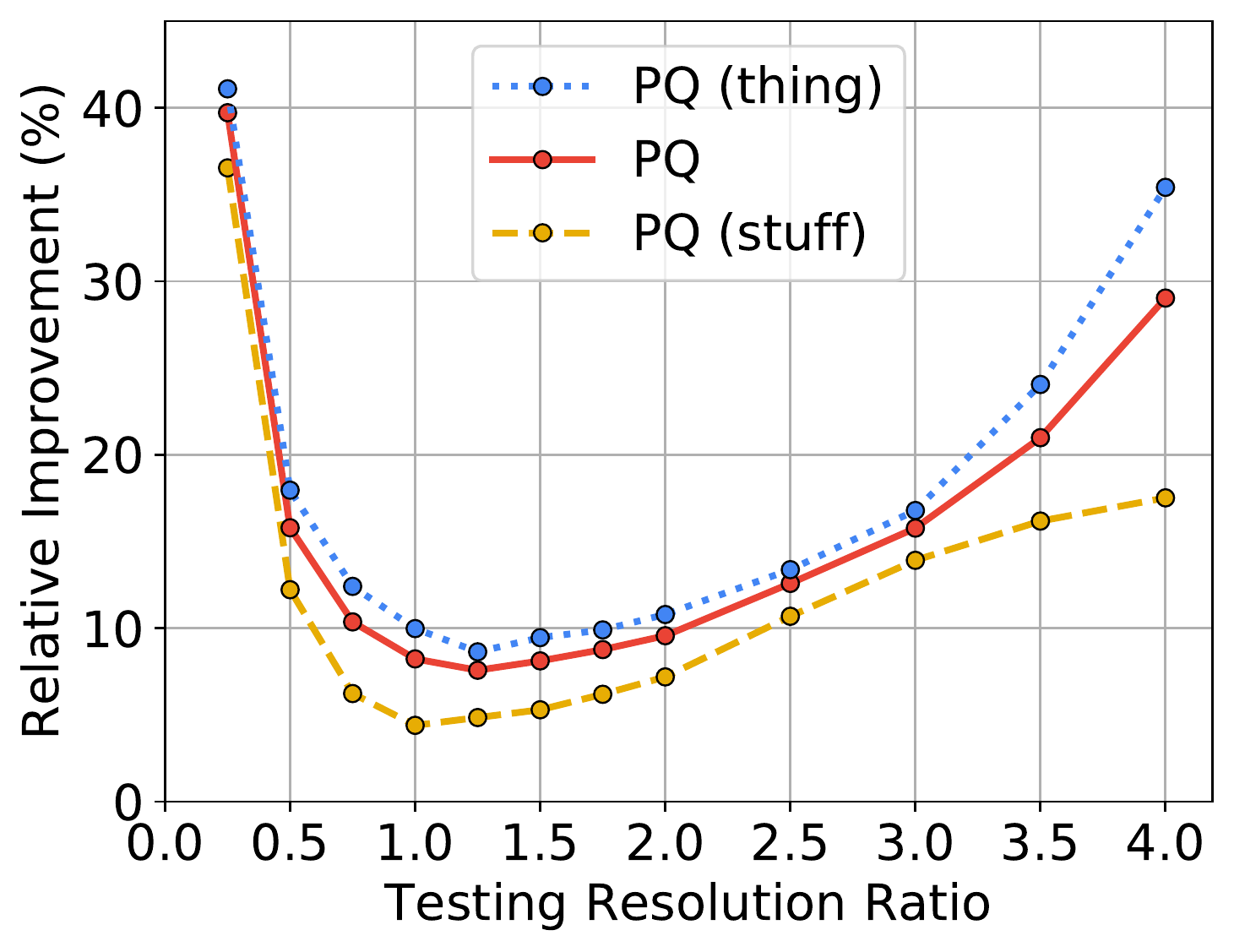}
    \caption{Scale stress test on COCO val set. Axial-DeepLab gains the most when tested on extreme resolutions. On the x-axis, ratio 4.0 means inference with resolution $4097\times4097$}
    \label{fig:scale}
\end{figure}

{\bf Scale Stress Test:} In order to verify that our model learns long range interactions, we perform a scale stress test besides standard testing. In the stress test, we train Panoptic-DeepLab (X-71) and our Axial-DeepLab-L with the standard setting, but test them on out-of-distribution resolutions (\ie, resize the input to different resolutions). \figref{fig:scale} summarizes our relative improvements over Panoptic-DeepLab on PQ, PQ (thing) and PQ (stuff). When tested on huge images, Axial-DeepLab shows large gain (30\%), demonstrating that it encodes long range relations better than convolutions. Besides, Axial-DeepLab improves 40\% on small images, showing that axial-attention is more robust to scale variations.

\subsection{Mapillary Vistas}

We evaluate our Axial-DeepLab on the large-scale Mapillary Vistas dataset \cite{neuhold2017mapillary}. We only report validation set results, since the test server is not available.

{\bf Val set:} As shown in \tabref{tab:mv_val}, our Axial-DeepLab-L outperforms all the state-of-the-art methods in both single-scale and multi-scale cases. Our {\it single-scale} Axial-DeepLab-L performs 2.4\% PQ better than the previous best {\it single-scale} Panoptic-DeepLab (X-71) \cite{cheng2019panoptic}. In multi-scale setting, our lightweight Axial-DeepLab-L performs better than Panoptic-DeepLab~(Auto-DeepLab-XL++), not only on panoptic segmentation (0.8\% PQ) and instance segmentation (0.3\% AP), but also on semantic segmentation (0.8\% mIoU), the task that Auto-DeepLab \cite{liu2019auto} was searched for. Additionally, to the best of our knowledge, our Axial-DeepLab-L attains the best {\it single-model} semantic segmentation result.

\begin{table}[!t]
\caption{Mapillary Vistas validation set. {\bf MS:} Multi-scale inputs}
\label{tab:mv_val}
\setlength{\tabcolsep}{0.2em}
    \centering
    \scalebox{0.87}{
    \begin{tabular}{l|c|cc|ccccc}
        \toprule[0.2em]
        Method & MS & Params & M-Adds & PQ & PQ\textsuperscript{Th} & PQ\textsuperscript{St} & AP & mIoU \\
        \toprule[0.2em]
        \multicolumn{9}{c}{Top-down panoptic segmentation methods} \\
        \midrule
        TASCNet \cite{li2018learning} & & & & 32.6 & 31.1 & 34.4 & 18.5 & - \\
        TASCNet \cite{li2018learning} & \ding{51} & & & 34.3 & 34.8 & 33.6 & 20.4 & - \\
        AdaptIS \cite{sofiiuk2019adaptis} & & & & 35.9 & 31.5 & 41.9 & - & - \\
        Seamless \cite{porzi2019seamless} & & & & 37.7 & 33.8 & 42.9 & 16.4 & 50.4 \\
        \midrule
        \multicolumn{9}{c}{Bottom-up panoptic segmentation methods} \\
        \midrule
        DeeperLab \cite{yang2019deeperlab} & & & & 32.0 & - & - & - & 55.3 \\
        Panoptic-DeepLab (Xception-71 \cite{chollet2016xception,dai2017coco}) \cite{cheng2019panoptic} & & 46.7M & 1.24T & 37.7 & 30.4 & 47.4 & 14.9 & 55.4 \\
        Panoptic-DeepLab (Xception-71 \cite{chollet2016xception,dai2017coco}) \cite{cheng2019panoptic} & \ding{51} & 46.7M & 31.35T & 40.3 & 33.5 & 49.3 & 17.2 & 56.8 \\
        Panoptic-DeepLab (HRNet-W48 \cite{wang2019deep}) \cite{cheng2019panoptic} & \ding{51} & 71.7M & 58.47T & 39.3 & - & - & 17.2 & 55.4 \\
        Panoptic-DeepLab (Auto-XL++ \cite{liu2019auto}) \cite{cheng2019panoptic} & \ding{51} & 72.2M & 60.55T & 40.3 & - & - & 16.9 & 57.6 \\
        \midrule \midrule
        Axial-DeepLab-L & & 44.9M & 1.55T & 40.1 & 32.7 & 49.8 & 16.7 & 57.6 \\
        Axial-DeepLab-L & \ding{51} & 44.9M & 39.35T & 41.1 & 33.4 & 51.3 & 17.2 & 58.4 \\
        \bottomrule[0.1em]
    \end{tabular}
    }
\end{table}

\subsection{Cityscapes}
\label{sec:cityscapes}

{\bf Val set:} In \tabref{tab:cityscapes_all}~(a), we report our Cityscapes validation set results. Without using extra data (\ie, only Cityscapes fine annotation), our Axial-DeepLab achieves 65.1\% PQ, which is 1\% better than the current best bottom-up Panoptic-DeepLab~\cite{cheng2019panoptic} and 3.1\% better than proposal-based AdaptIS \cite{sofiiuk2019adaptis}. When using extra data (\eg, Mapillary Vistas \cite{neuhold2017mapillary}), our {\it multi-scale} Axial-DeepLab-XL attains 68.5\% PQ, 1.5\% better than Panoptic-DeepLab~\cite{cheng2019panoptic} and 3.5\% better than Seamless~\cite{porzi2019seamless}. Our instance segmentation and semantic segmentation results are respectively 1.7\% and 1.5\% better than Panoptic-DeepLab~\cite{cheng2019panoptic}.

    \begin{table}[!t]
        \caption{Cityscapes val set and test set. {\bf MS:} Multi-scale inputs. {\bf C:} Cityscapes coarse annotation. {\bf V:} Cityscapes video. {\bf MV:} Mapillary Vistas
        }
        \setlength{\tabcolsep}{0.1em}
        \label{tab:cityscapes_all}
        \begin{tabular}{cc}
        (a) Cityscapes validation set & (b) Cityscapes test set \\
        \scalebox{0.77}{
        \begin{tabular}{l|c|c|ccc}
        \toprule[0.2em]
            Method & Extra Data & MS & PQ & AP & mIoU \\
            \toprule[0.2em]
            AdaptIS~\cite{sofiiuk2019adaptis} & & \cmark & 62.0 & 36.3 & 79.2 \\
            \midrule
            SSAP~\cite{gao2019ssap} & & \cmark & 61.1 & 37.3 & - \\
            Panoptic-DeepLab~\cite{cheng2019panoptic} & & & 63.0 & 35.3 & 80.5 \\
            Panoptic-DeepLab~\cite{cheng2019panoptic} & & \cmark & 64.1 & 38.5 & {\bf 81.5} \\
            \midrule
            Axial-DeepLab-L & & & 63.9 & 35.8 & 81.0 \\
            Axial-DeepLab-L & & \cmark & 64.7 & 37.9 & {\bf 81.5} \\
            Axial-DeepLab-XL & & & 64.4 & 36.7 & 80.6 \\
            Axial-DeepLab-XL & & \cmark & {\bf 65.1} & {\bf 39.0} & 81.1 \\
            \midrule \midrule
            SpatialFlow~\cite{chen2019spatialflow} & COCO & \cmark & 62.5 & - & - \\
            Seamless~\cite{porzi2019seamless} & MV &  & 65.0 & - & 80.7 \\
            \midrule
            Panoptic-DeepLab~\cite{cheng2019panoptic} & MV & & 65.3 & 38.8 & 82.5 \\
            Panoptic-DeepLab~\cite{cheng2019panoptic} & MV & \cmark & 67.0 & 42.5 & 83.1 \\
            \midrule
            Axial-DeepLab-L & MV & & 66.5 & 40.2 & 83.2 \\
            Axial-DeepLab-L & MV & \cmark & 67.7 & 42.9 & 83.8\\
            Axial-DeepLab-XL & MV & & 67.8 & 41.9 & 84.2 \\
            Axial-DeepLab-XL & MV & \cmark & {\bf 68.5} & {\bf 44.2} & {\bf 84.6}\\
            \bottomrule[0.1em]
        \end{tabular}
        } & 
        \scalebox{0.77}{
        \begin{tabular}{l|c|ccc}
        \toprule[0.2em]
            Method & Extra Data & PQ & AP & mIoU \\
            \toprule[0.2em]
            GFF-Net~\cite{li2019gff} & & - & - & 82.3 \\
            Zhu~\etal~\cite{zhu2019improving} & C, V, MV & - & - & 83.5 \\
            \midrule
            AdaptIS~\cite{sofiiuk2019adaptis} & & - & 32.5 & - \\
            UPSNet~\cite{xiong2019upsnet} & COCO & - & 33.0 & - \\
            PANet~\cite{liu2018path} & COCO & - & 36.4 & - \\
            PolyTransform~\cite{liang2019polytransform} & COCO & - & 40.1 & \\
            \midrule
            SSAP~\cite{gao2019ssap} & & 58.9 & 32.7 & - \\
            Li~\etal\cite{li2020unifying} &  & 61.0 & - & - \\
            Panoptic-DeepLab~\cite{cheng2019panoptic} & & 62.3 & 34.6 & 79.4 \\
            TASCNet~\cite{li2018learning} & COCO & 60.7 & - & - \\
            Seamless~\cite{porzi2019seamless} & MV & 62.6 & - & - \\
            Li~\etal\cite{li2020unifying} & COCO & 63.3 & - & - \\
            Panoptic-DeepLab~\cite{cheng2019panoptic} & MV & 65.5 & 39.0 & 84.2 \\
            \midrule \midrule
            Axial-DeepLab-L & & 62.7 & 33.3 & 79.5 \\
            Axial-DeepLab-XL & & 62.8 & 34.0 & 79.9 \\
            Axial-DeepLab-L & MV & 65.6 & 38.1 & 83.1 \\
            Axial-DeepLab-XL & MV & {\bf 66.6} & 39.6 & 84.1 \\
            \bottomrule[0.1em]
        \end{tabular}
        } \\
        \end{tabular}
\end{table}

{\bf Test set:} \tabref{tab:cityscapes_all}~(b) shows our test set results. Without extra data, Axial-DeepLab-XL attains 62.8\% PQ, setting a new state-of-the-art result. Our model further achieves 66.6\% PQ, 39.6\% AP, and 84.1\% mIoU with Mapillary Vistas pretraining. Note that  Panoptic-DeepLab~\cite{cheng2019panoptic} adopts the trick of output stride 8 during inference on test set, making their M-Adds comparable to our XL models.

\subsection{Ablation Studies}

We perform ablation studies on Cityscapes validation set.

{\bf Importance of Position-Sensitivity and Axial-Attention:} In \tabref{tab:imagenet}, we experiment with attention models on ImageNet. In this ablation study, we transfer them to Cityscapes segmentation tasks. As shown in \tabref{tab:ablation1}, all variants outperform ResNet-50 \cite{he2016deep}. Position-sensitive attention performs better than previous self-attention \cite{parmar2019stand}, which aligns with ImageNet results in \tabref{tab:imagenet}. However, employing axial-attention, which is on-par with position-sensitive attention on ImageNet, gives more than 1\% boosts on all three segmentation tasks (in PQ, AP, and mIoU), without ASPP, and with fewer parameters and M-Adds, suggesting that the ability to encode long range context of axial-attention significantly improves the performance on segmentation tasks with large input images. 

    \begin{table}[!t]
    \caption{Ablating self-attention variants on Cityscapes val set. {\bf ASPP}: Atrous spatial pyramid pooling. {\bf PS}: Our position-sensitive self-attention}
    \label{tab:ablation1}
    \setlength{\tabcolsep}{0.5em}
        \centering
        \scalebox{1.0}{
        \begin{tabular}{l|c|c|cc|ccc}
        \toprule[0.2em]
            Backbone & ASPP & PS & Params & M-Adds & PQ & AP & mIoU \\
            \toprule[0.2em]
            ResNet-50~\cite{he2016deep} (our impl.) & & & 24.8M & 374.8B & 58.1 & 30.0 & 73.3 \\
            ResNet-50~\cite{he2016deep} (our impl.) & \ding{51} & & 30.0M & 390.0B & 59.8 & 32.6 & 77.8\\
            Attention~\cite{parmar2019stand} (our impl.) & & & 17.3M & 317.7B & 58.7 & 31.9 & 75.8 \\
            Attention~\cite{parmar2019stand} (our impl.) & \ding{51} & & 22.5M & 332.9B & 60.9 & 30.0 & 78.2 \\
            \midrule
            PS-Attention & & \ding{51} & 17.3M & 326.7B & 59.9 & 32.2 & 76.3 \\
            PS-Attention & \ding{51} & \ding{51} & 22.5M & 341.9B & \bf{61.5} & \bf{33.1} & \bf{79.1} \\
            \midrule
            Axial-DeepLab-S & & \ding{51} & \bf{12.1M} & \bf{220.8B} & \bf{62.6} & \bf{34.9} & \bf{80.5} \\
            \midrule
            Axial-DeepLab-M & & \ding{51} & 25.9M & 419.6B & 63.1 & 35.6 & 80.3 \\
            Axial-DeepLab-L & & \ding{51} & 44.9M & 687.4B & 63.9 & 35.8 & 81.0 \\
            Axial-DeepLab-XL & & \ding{51} & 173.0M & 2446.8B & 64.4 & 36.7 & 80.6 \\
            \bottomrule[0.1em]
        \end{tabular}
        }
    \end{table}

{\bf Importance of Axial-Attention Span:} In \tabref{tab:ablation2}, we vary the span $m$ (\ie, spatial extent of local regions in an axial block), without ASPP. We observe that a larger span consistently improves the performance at marginal costs.

    \begin{table}[!t]
    \caption{Varying axial-attention span on Cityscapes val set}
    \label{tab:ablation2}
    \setlength{\tabcolsep}{0.7em}
        \centering
        \scalebox{1.0}{
        \begin{tabular}{l|c|cc|ccc}
        \toprule[0.2em]
            Backbone & Span & Params & M-Adds & PQ & AP & mIoU \\
            \toprule[0.2em]
            ResNet-101 & - & 43.8M & 530.0B & 59.9 & 31.9 & 74.6 \\
            \midrule
            Axial-ResNet-L & 5 $\times$ 5 & 44.9M & 617.4B & 59.1 & 31.3 & 74.5 \\
            Axial-ResNet-L & 9 $\times$ 9 & 44.9M & 622.1B & 61.2 & 31.1 & 77.6 \\
            Axial-ResNet-L & 17 $\times$ 17 & 44.9M & 631.5B & 62.8 & 34.0 & 79.5 \\
            Axial-ResNet-L & 33 $\times$ 33 & 44.9M & 650.2B & 63.8 & 35.9 & 80.2 \\
            Axial-ResNet-L & 65 $\times$ 65 & 44.9M & 687.4B & \bf{64.2} & \bf{36.3} & \bf{80.6} \\
            \bottomrule[0.1em]
        \end{tabular}
        }
    \end{table}

\section{Conclusion and Discussion}
\label{sec:discussion}

In this work, we have shown the effectiveness of proposed position-sensitive axial-attention on image classification and segmentation tasks. On ImageNet, our Axial-ResNet, formed by stacking axial-attention blocks, achieves state-of-the-art results among stand-alone self-attention models. We further convert Axial-ResNet to Axial-DeepLab for bottom-up segmentation tasks, and also show state-of-the-art performance on several benchmarks, including COCO, Mapillary Vistas, and Cityscapes. We hope our promising results could establish that axial-attention is an effective building block for modern computer vision models.

Our method bears a similarity to decoupled convolution~\cite{DBLP:conf/bmvc/JaderbergVZ14}, which factorizes a depthwise convolution~\cite{sifre2014rigid,howard2017mobilenets,chollet2016xception} to a column convolution and a row convolution. This operation could also theoretically achieve a large receptive field, but its convolutional template matching nature limits the capacity of modeling multi-scale interactions. Another related method is deformable convolution~\cite{dai2017deformable,zhu2019deformable,gao2019deformable}, where each point attends to a few points dynamically on an image. However, deformable convolution does not make use of key-dependent positional bias or content-based relation. In addition, axial-attention propagates information densely, and more efficiently along the height- and width-axis sequentially.

Although our axial-attention model saves M-Adds, it runs slower than convolutional counterparts, as also observed by \cite{parmar2019stand}. This is due to the lack of specialized kernels on various accelerators for the time being. This might well be improved if the community considers axial-attention as a plausible direction.

\section*{Acknowledgments}
We thank Niki Parmar for discussion and support; Ashish Vaswani, Xuhui Jia, Raviteja Vemulapalli, Zhuoran Shen for their insightful comments and suggestions; Maxwell Collins and Blake Hechtman for technical support. This work is supported by Google Faculty Research Award and NSF 1763705.

\newpage
\begin{subappendices}
\renewcommand{\thesection}{\Alph{section}}%
\section{Runtime}
In this section, we profile our Conv-Stem Axial-ResNet-L in a common setting: 224x224 feed-forward with batch size 1, on a V100 GPU, averaged over 5 runs. The time includes input standardization, and the last projection to 1000 logits. Our model takes 16.54 ms. For comparison, we list our TensorFlow runs of some popular models at hand (with comparable flops). To provide more context, we take entries from \cite{bianco2018benchmark} for reference (A Titan X Pascal is used in \cite{bianco2018benchmark}, but the PyTorch code is more optimized). Our runtime is roughly at the same level of ResNeXt-101 (32x4d), SE-ResNet-101, ResNet-152, and DenseNet-201 (k=32).

Note that we directly benchmark with our code optimized for TPU execution, with channels being the last dimension. Empirically, the generated graph involves transposing between NCHW and NHWC, before and after almost every conv2d operation. (This effect also puts Xception-71 at a disadvantage because of its separable conv design.) Further optimizing this could lead to faster inference.

    \begin{table}[t]
    \caption{Runtime of Axial-ResNet-L on a 224$\times$224 image}
    \label{tab:runtime}
    \setlength{\tabcolsep}{1.5em}
        \centering
        \begin{tabular}{l|c|c}
        \toprule[0.2em]
            Model & Our Profile (ms) & \cite{bianco2018benchmark} (ms) \\
            \toprule[0.2em]
            Axial-ResNet-L & 16.54 & - \\
            \midrule
            Stand-Alone-L~\cite{parmar2019stand} & 18.05 & - \\
            Xception-71~\cite{chollet2016xception,deeplabv3plus2018} & 24.85 & - \\
            ResNet-101~\cite{he2016deep} & 10.08 & 8.9 \\
            ResNet-152~\cite{he2016deep} & 14.43 & 14.31 \\
            ResNeXt-101 (32x4d)~\cite{xie2017aggregated} & - & 17.05 \\
            SE-ResNet-101~\cite{hu2018squeeze} & - & 15.10 \\
            SE-ResNeXt-101 (32x4d)~\cite{hu2018squeeze} & - & 24.96 \\
            DenseNet-201 (k=32)~\cite{huang2017densely} & - & 17.15 \\
            \bottomrule[0.1em]
        \end{tabular}
    \end{table}
    
We observe that our Conv-Stem Axial-ResNet-L runs faster than Conv-Stem Stand-Alone-L~\cite{parmar2019stand}, although we split one layer into two. This is because our axial-attention makes better use of existing kernels:
\begin{itemize}
  \item The width-axis attention is parallelizable over height-axis, i.e. this is a large batch of 1d row operations (the batch size is the height of the input).
  \item Axial attention avoids extracting 2d memory blocks with pads, splits and concatenations, which are not efficient on accelerators.
\end{itemize}

\section{Axial-Decoder}
\begin{figure}[t]
    \centering
    \includegraphics[width=0.99\textwidth]{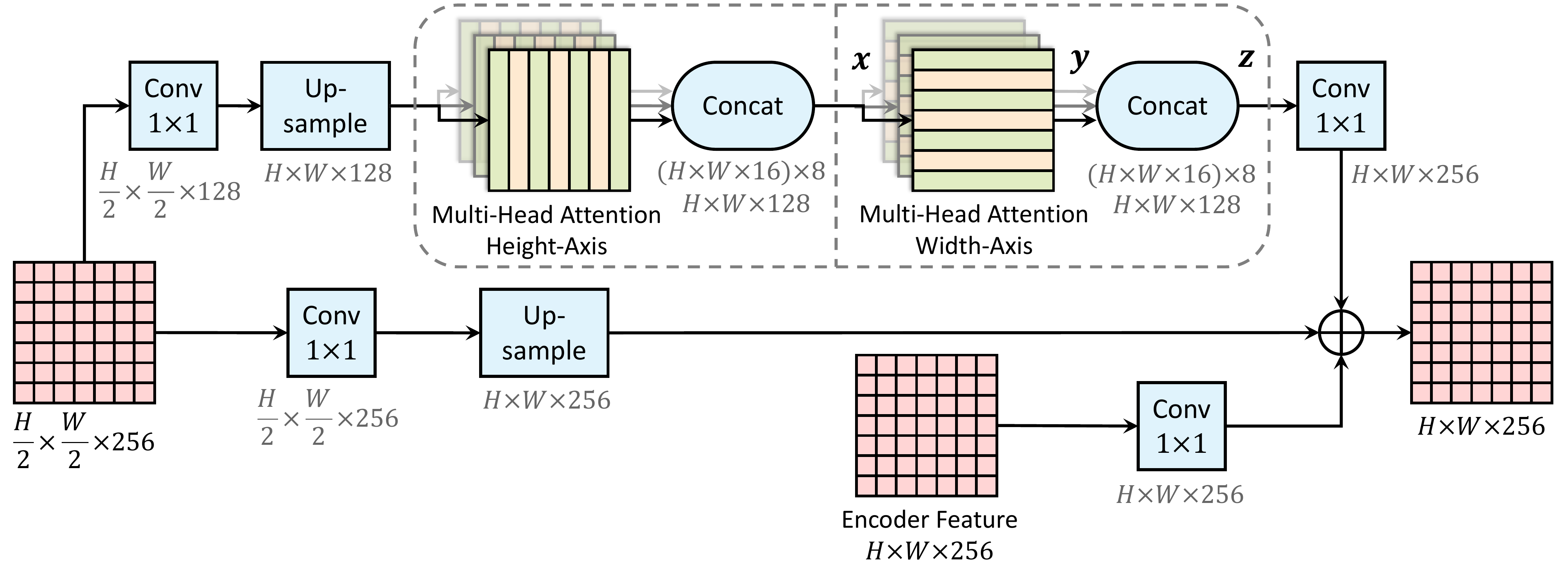}
    \caption{An axial-decoder block. We augment an axial-attention block with up-samplings, and encoder features}
    \label{fig:axial_decoder}
\end{figure}

Axial-DeepLab employs dual convolutional decoders~\cite{cheng2019panoptic}. In this section, we explore a setting with a {\it single} axial-decoder instead. In the axial-decoder module, we apply one axial-attention block at each upsampling stage. In \figref{fig:axial_decoder}, we show an example axial-decoder in Axial-DeepLab-L from output stride 8 to output stride 4. We apply three such blocks, analogous to the three 5$\times$5 convolutions in Panoptic-DeepLab~\cite{cheng2019panoptic}.

{\bf Importance of Output Stride and Axial-Decoder:} In \tabref{tab:ablation3}, we experiment with the effect of output stride and axial-decoder (\ie, replacing dual decoders with axial-attention blocks). As shown in the table, our models are robust to output stride, and using axial-decoder is able to yield similar results. Our simple axial-decoder design works as well as dual convolutional decoders.

    \begin{table}[t]
    \caption{Ablating output strides and decoder types on Cityscapes val set. {\bf ASPP}: Atrous spatial pyramid pooling. {\bf OS}: Output stride (\ie, the ratio of image resolution to final feature resolution in backbone). {\bf AD}: Use axial-decoder in Axial-DeepLab}
    \label{tab:ablation3}
    \setlength{\tabcolsep}{0.5em}
        \centering
        \scalebox{1.0}{
        \begin{tabular}{l|c|c|c|cc|ccc}
        \toprule[0.2em]
            Backbone & ASPP & OS & AD & Params & M-Adds & PQ & AP & mIoU \\
            \toprule[0.2em]
            Xception-71 & \ding{51} & 16 & & 46.7M & 547.7B & 63.2 & 35.0 & 80.2\\
            \midrule
            Axial-ResNet-L & & 16 & & 44.9M & 687.4B & 63.9 & 35.8 & 81.0 \\
            Axial-ResNet-L & & 32 & & 45.2M & 525.2B & 63.9 & 36.3 & 80.9 \\
            Axial-ResNet-L & & 16 & \ding{51} & 45.4M & 722.7B & 63.7 & 36.9 & 80.7 \\
            Axial-ResNet-L & & 32 & \ding{51} & 45.9M & 577.8B & 64.0 & 37.1 & 81.0 \\
            \bottomrule[0.1em]
        \end{tabular}
        }
    \end{table}

\section{COCO Visualization}
\begin{figure}
    \centering
    \setlength\tabcolsep{2.12345pt}
    \begin{tabular}{c|c|c|c}
        \includegraphics[width=0.23\textwidth]{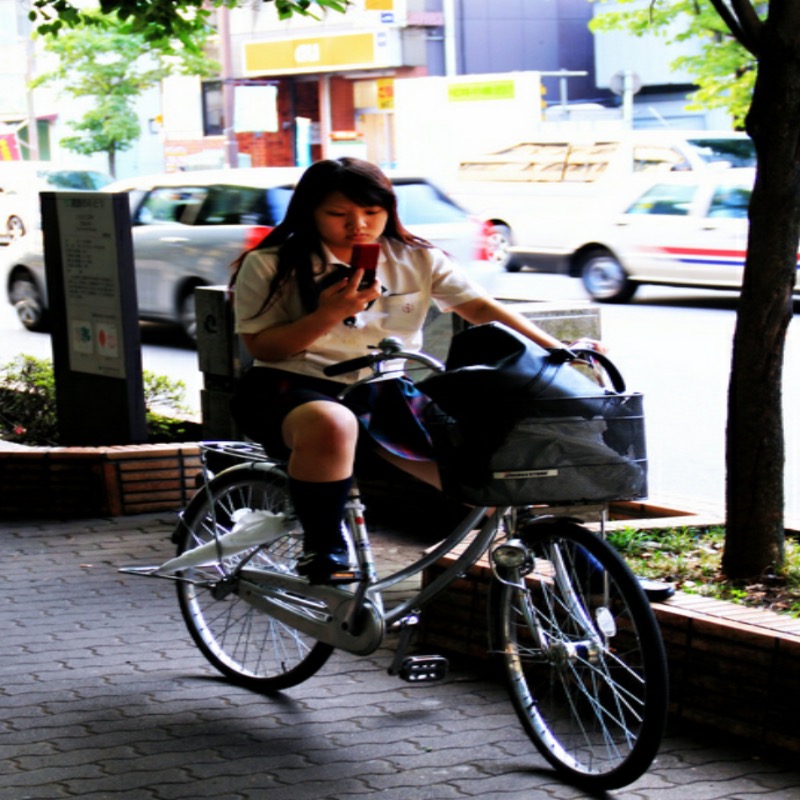} &
        \includegraphics[width=0.23\textwidth]{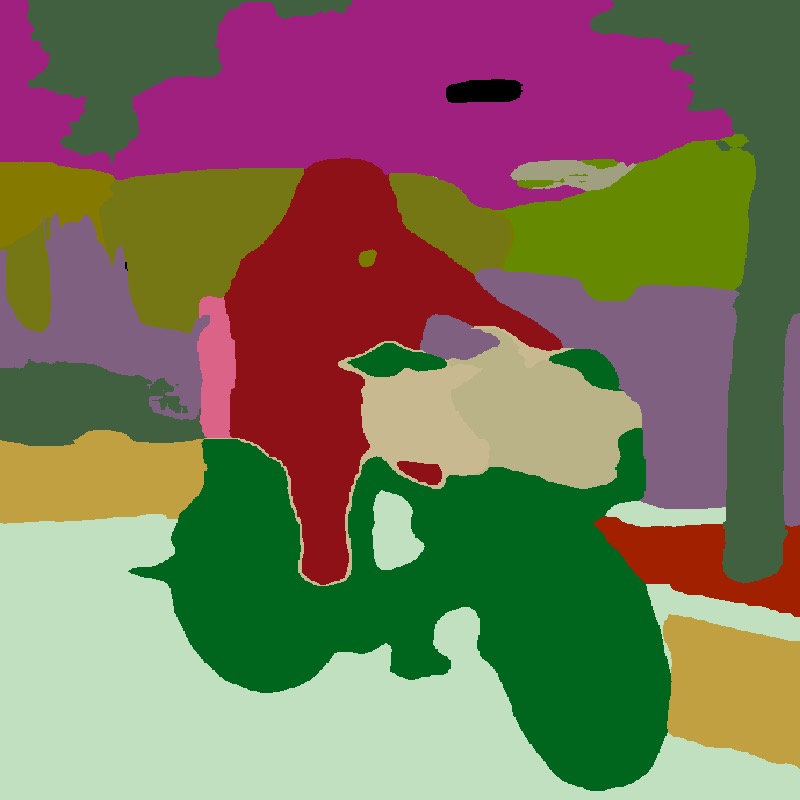} &
        \includegraphics[width=0.23\textwidth]{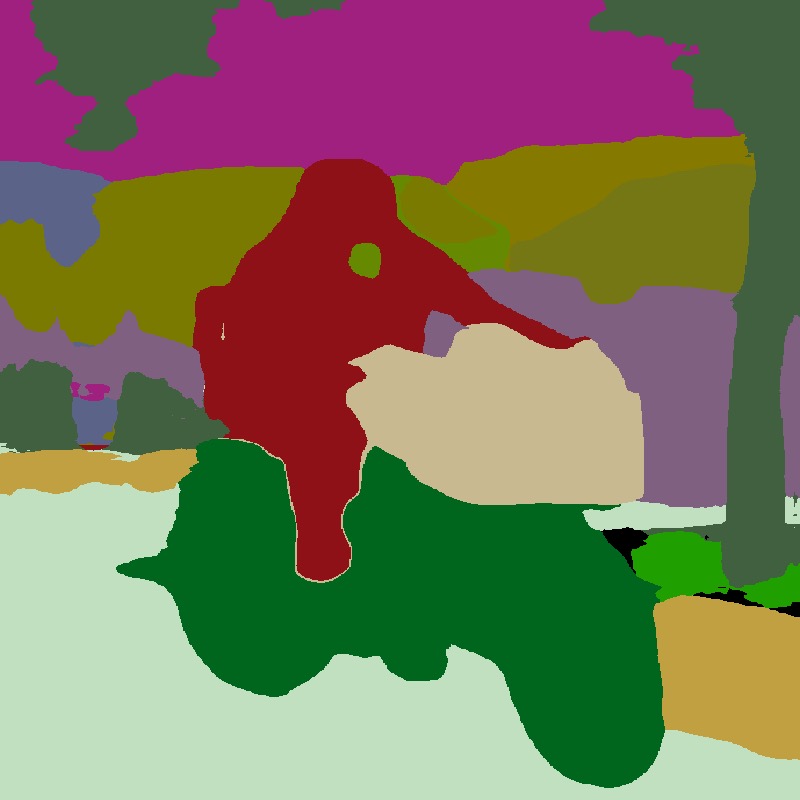} &
        \includegraphics[width=0.23\textwidth]{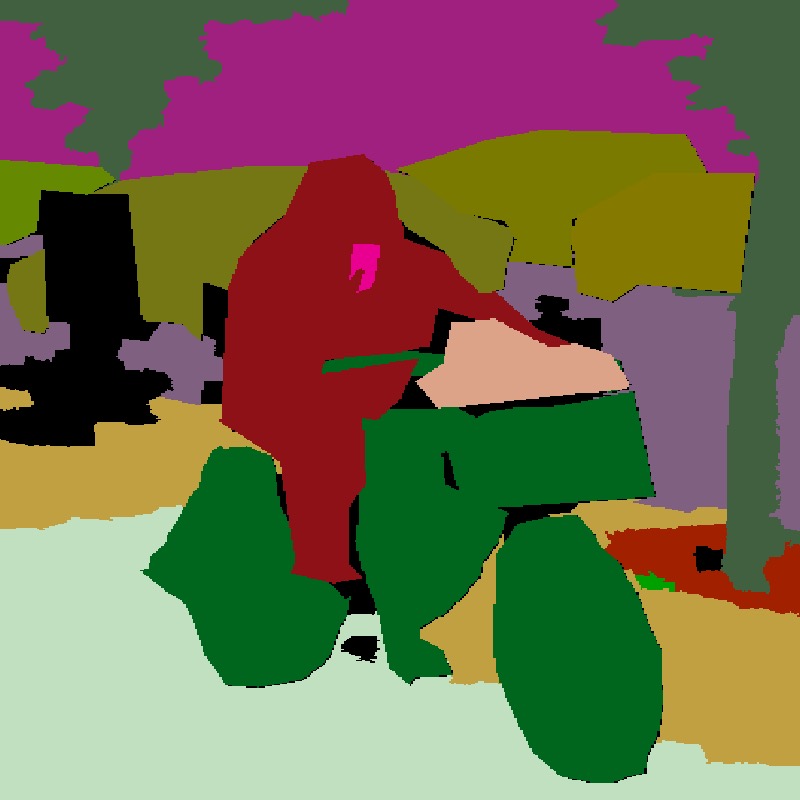} \\
        \includegraphics[width=0.23\textwidth]{figures_appendix/000025_image.jpg} &
        \includegraphics[width=0.23\textwidth]{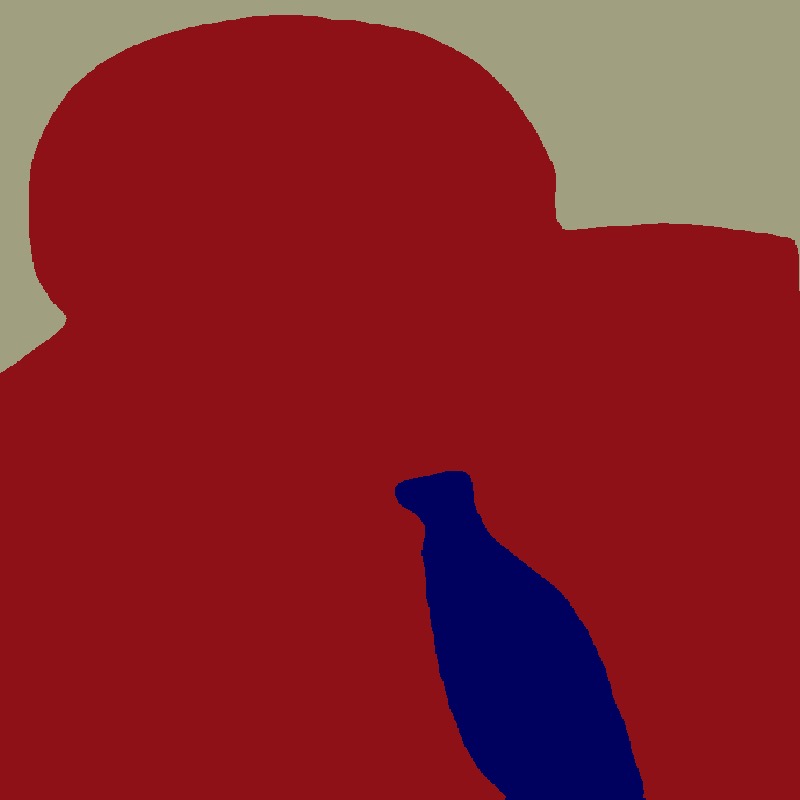} &
        \includegraphics[width=0.23\textwidth]{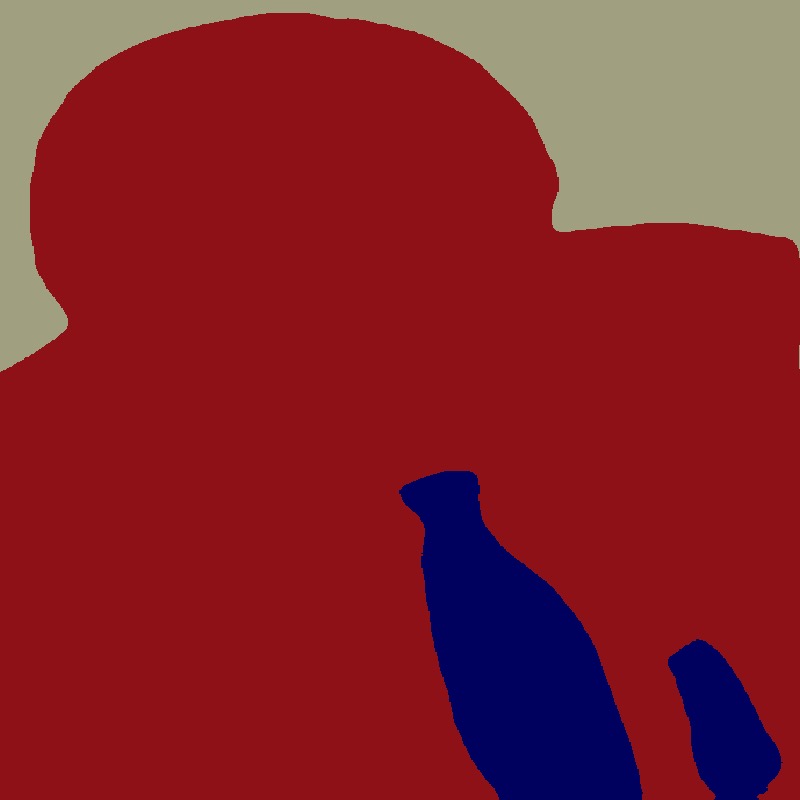} &
        \includegraphics[width=0.23\textwidth]{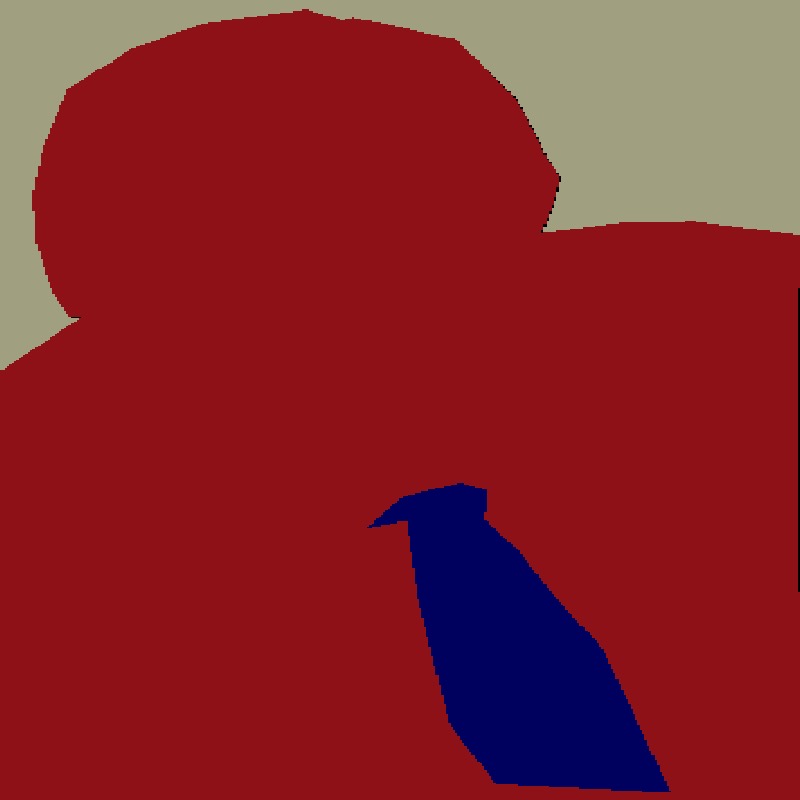} \\
        \includegraphics[width=0.23\textwidth]{figures_appendix/000042_image.jpg} &
        \includegraphics[width=0.23\textwidth]{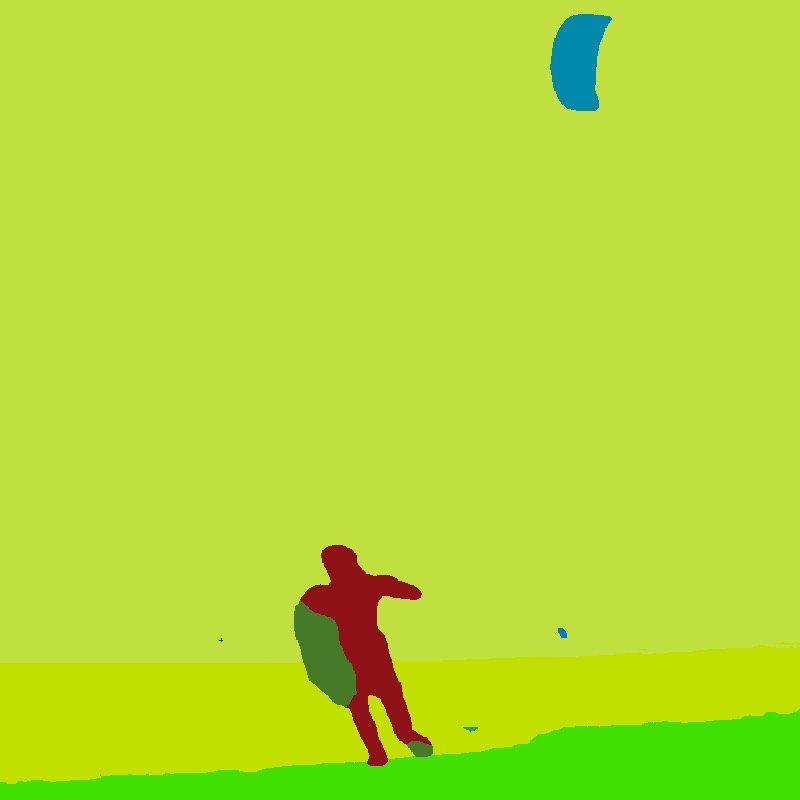} &
        \includegraphics[width=0.23\textwidth]{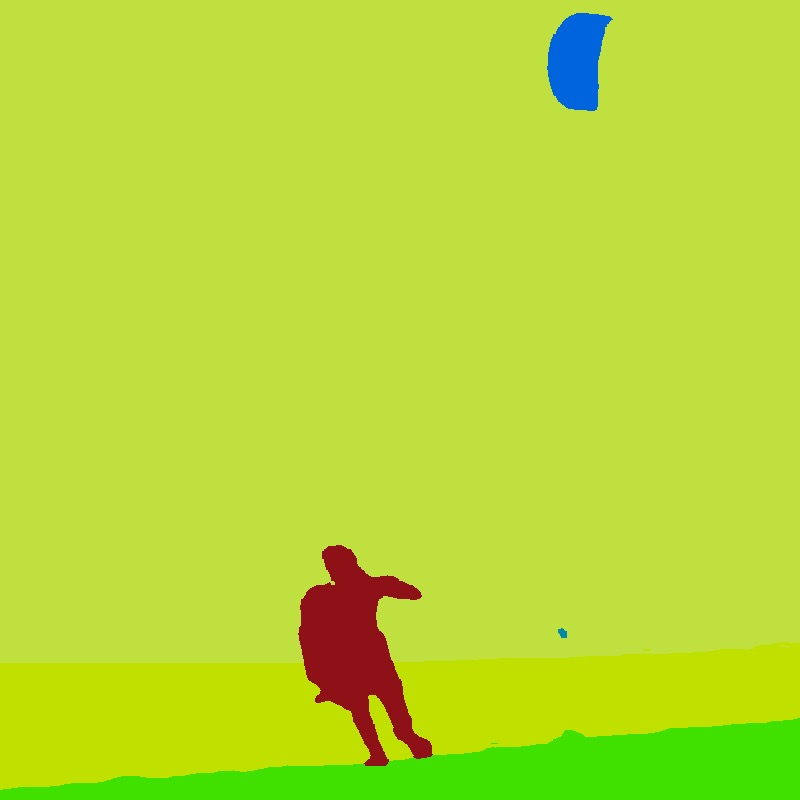} &
        \includegraphics[width=0.23\textwidth]{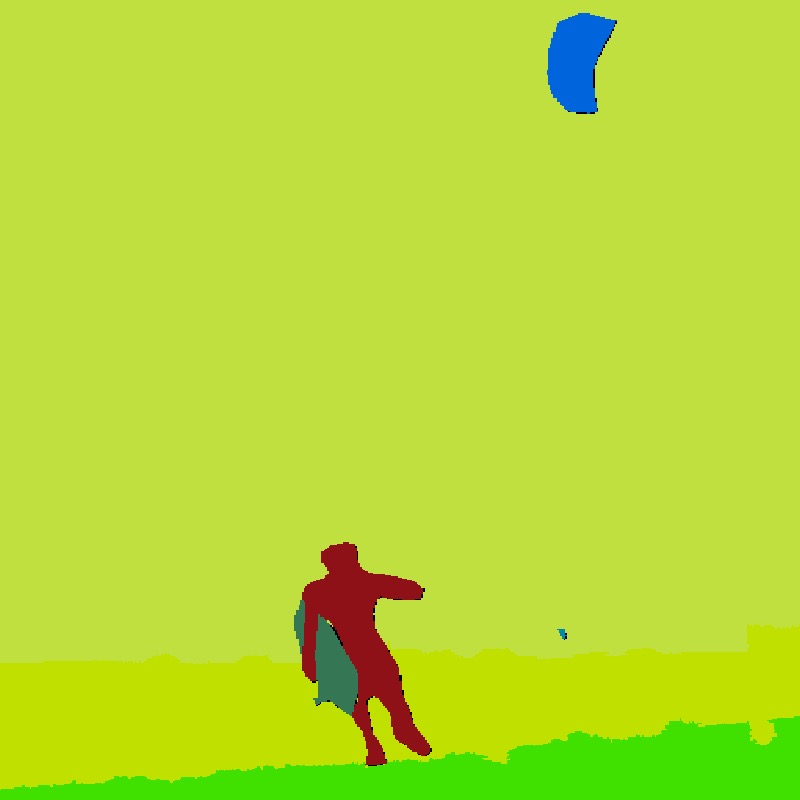} \\
        \includegraphics[width=0.23\textwidth]{figures_appendix/000064_image.jpg} &
        \includegraphics[width=0.23\textwidth]{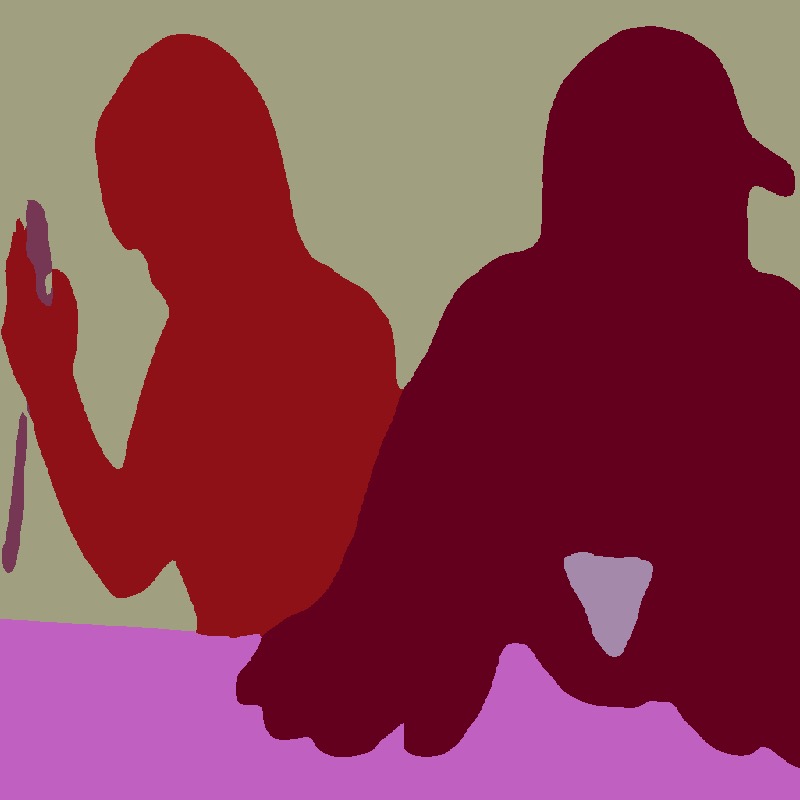} &
        \includegraphics[width=0.23\textwidth]{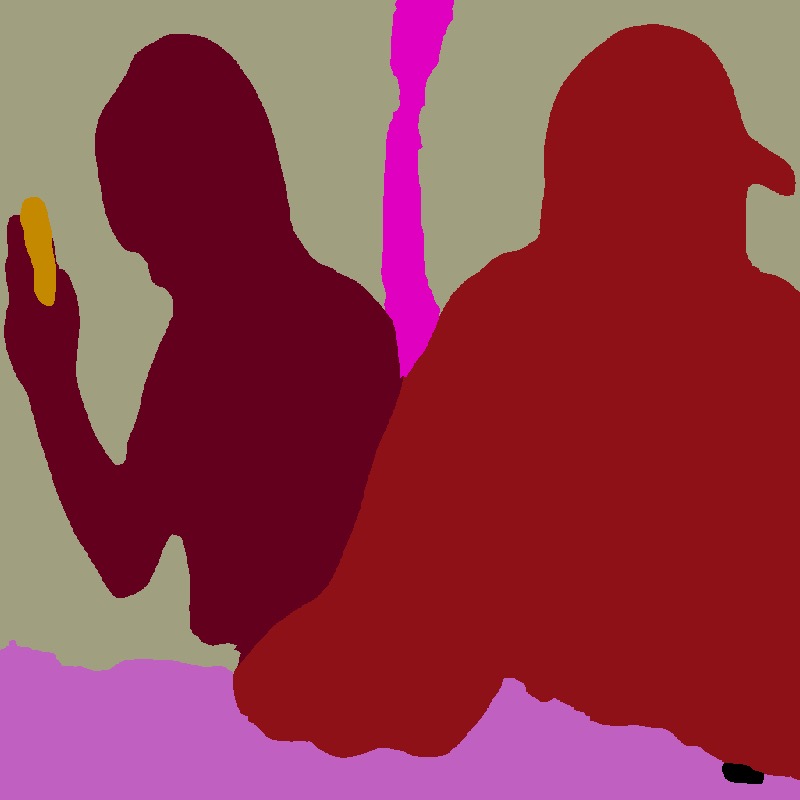} &
        \includegraphics[width=0.23\textwidth]{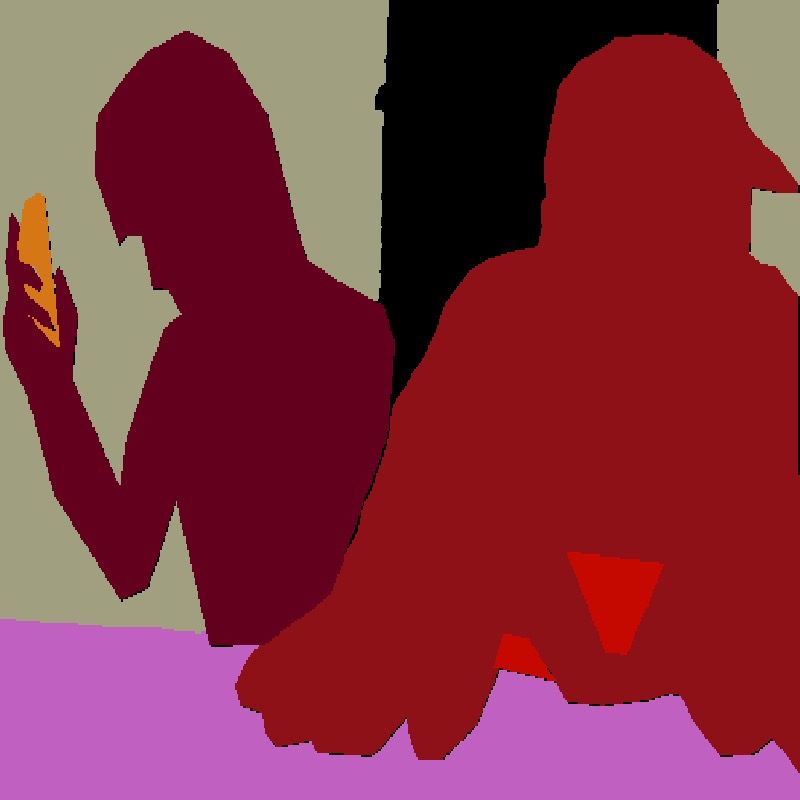} \\
        \includegraphics[width=0.23\textwidth]{figures_appendix/000086_image.jpg} &
        \includegraphics[width=0.23\textwidth]{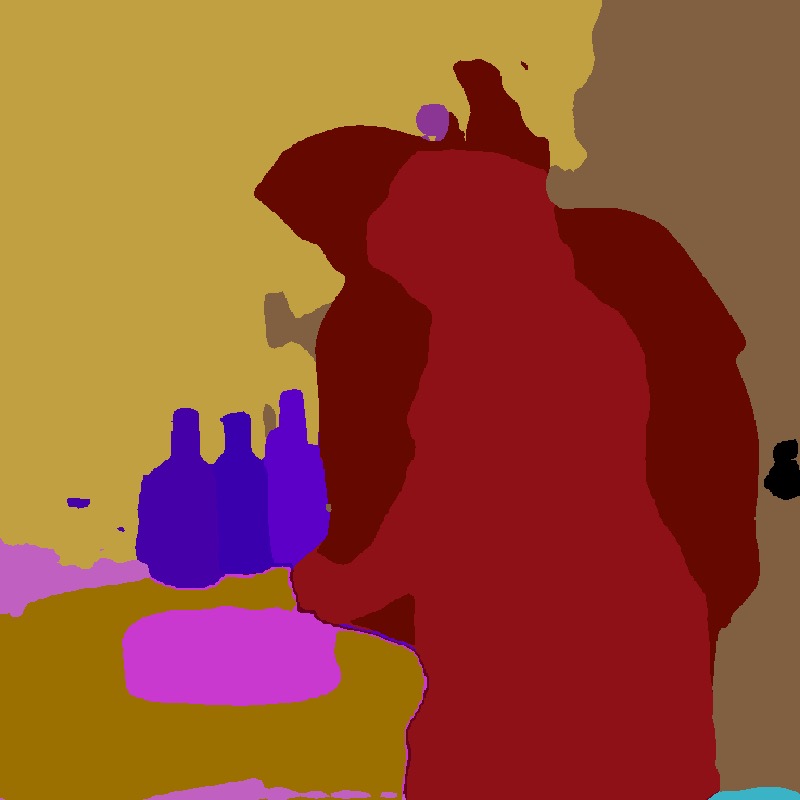} &
        \includegraphics[width=0.23\textwidth]{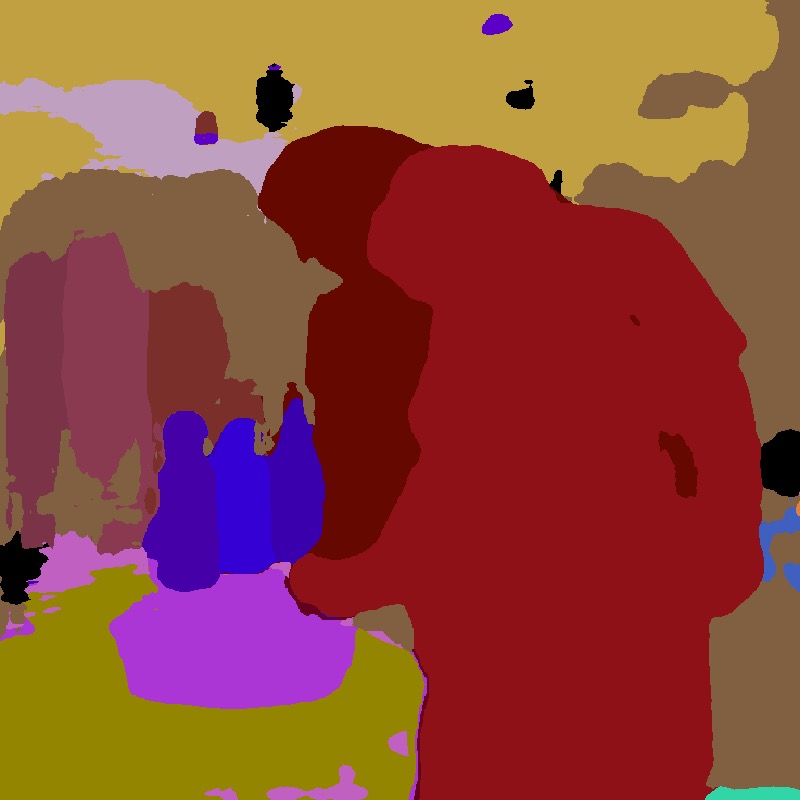} &
        \includegraphics[width=0.23\textwidth]{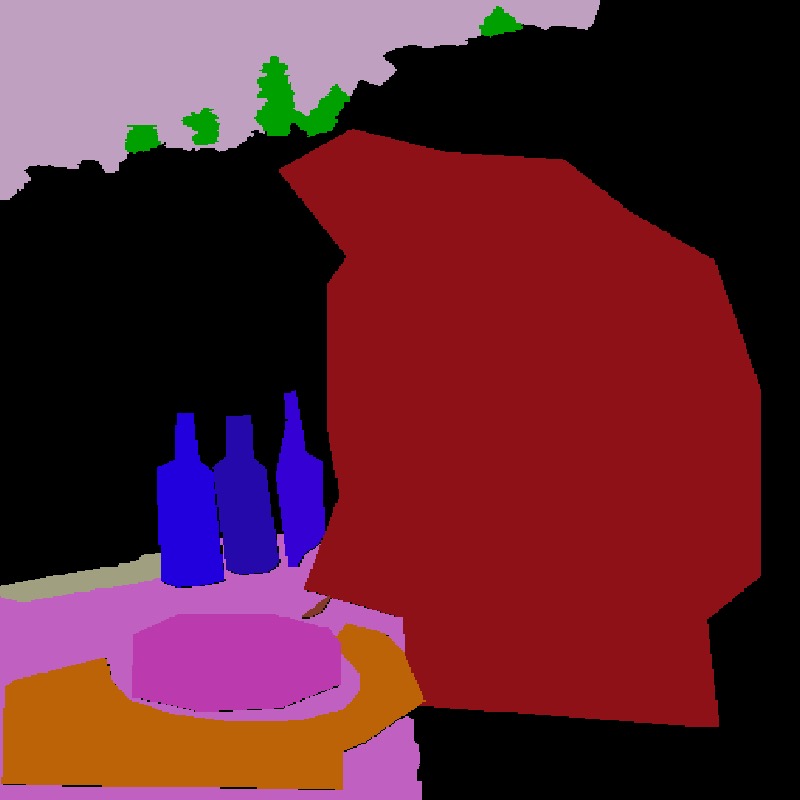} \\
        Original Image & Axial-DeepLab & Panoptic-DeepLab & Ground Truth \\
    \end{tabular}
    \caption{Visualization on COCO val set. Axial-DeepLab shows robustness to occlusion. In row 1 and row 4, Axial-DeepLab captures the occluded left leg and the remote control cable respectively, which are not even present in ground truth labels. In the last row, Axial-DeepLab distinguishes one person occluding another correctly, whereas the ground truth treats them as one instance}
    \label{fig:visualization}
\end{figure}

In \figref{fig:visualization}, we visualize some panoptic segmentation results on COCO val set. Our Axial-DeepLab-L demonstrates robustness to occlusion, compared with Panoptic-DeepLab (Xception-71).

\begin{figure}
    \centering
    \setlength\tabcolsep{3.12345pt}
    \begin{tabular}{cccc}
    \multicolumn{2}{c}{Original Image} & \multicolumn{2}{c}{Panoptic Prediction} \\
    \multicolumn{2}{c}{\includegraphics[width=0.35\textwidth]{figures_appendix/000044_image.jpg}} & \multicolumn{2}{c}{\includegraphics[width=0.35\textwidth]{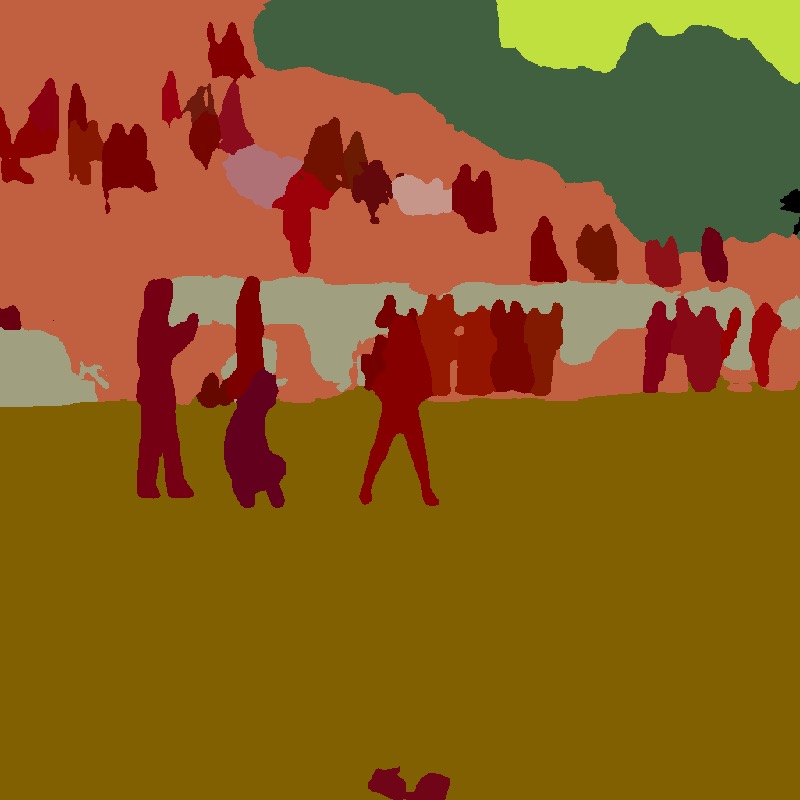}} \\
    column head 1 & column head 2 & column head 3 & column head 4 \\
    \includegraphics[width=0.2000\textwidth]{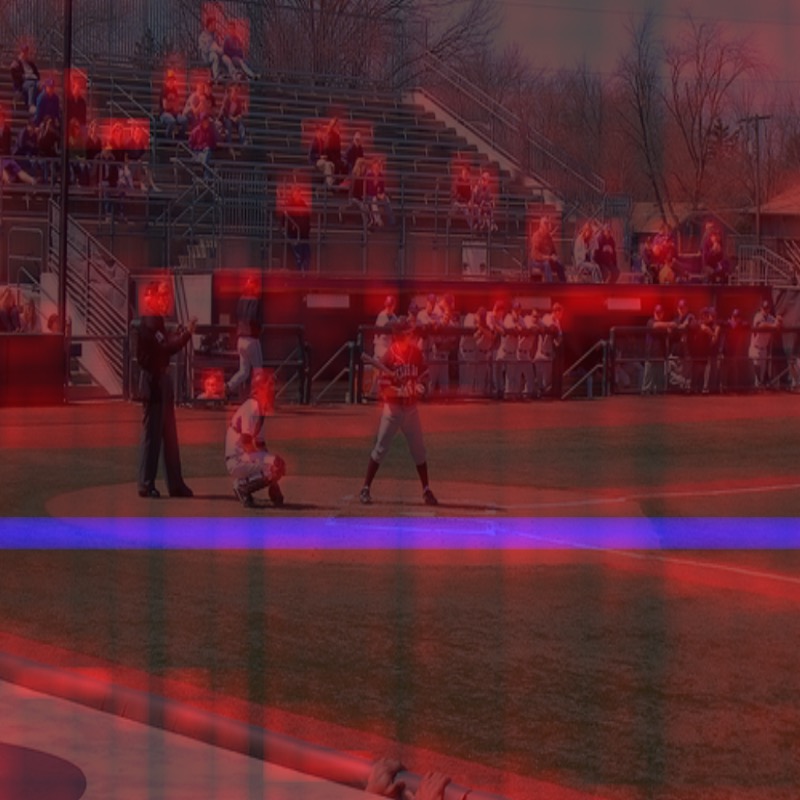} & \includegraphics[width=0.2000\textwidth]{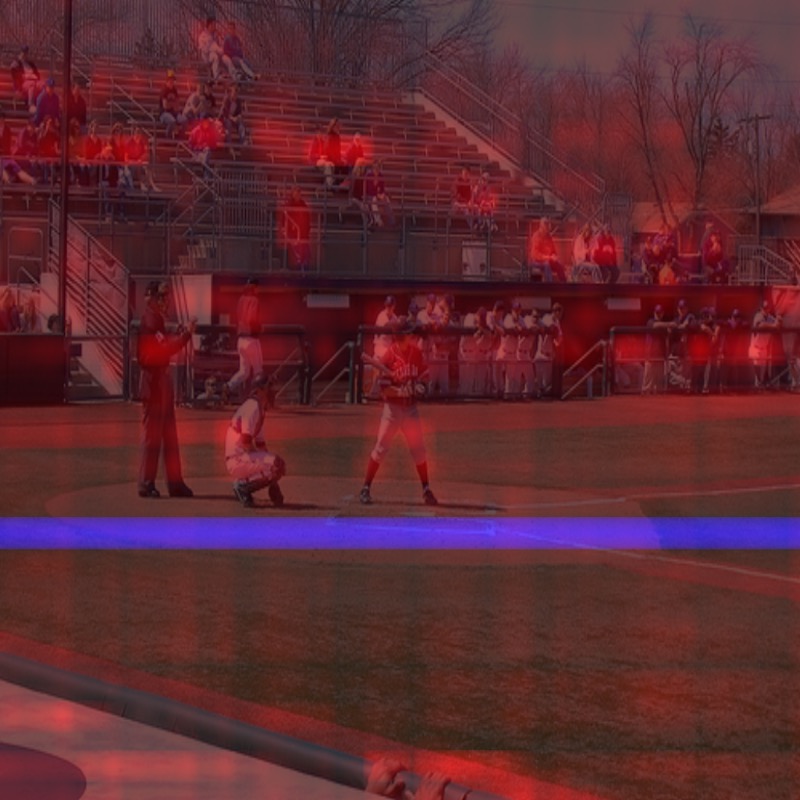} & \includegraphics[width=0.2000\textwidth]{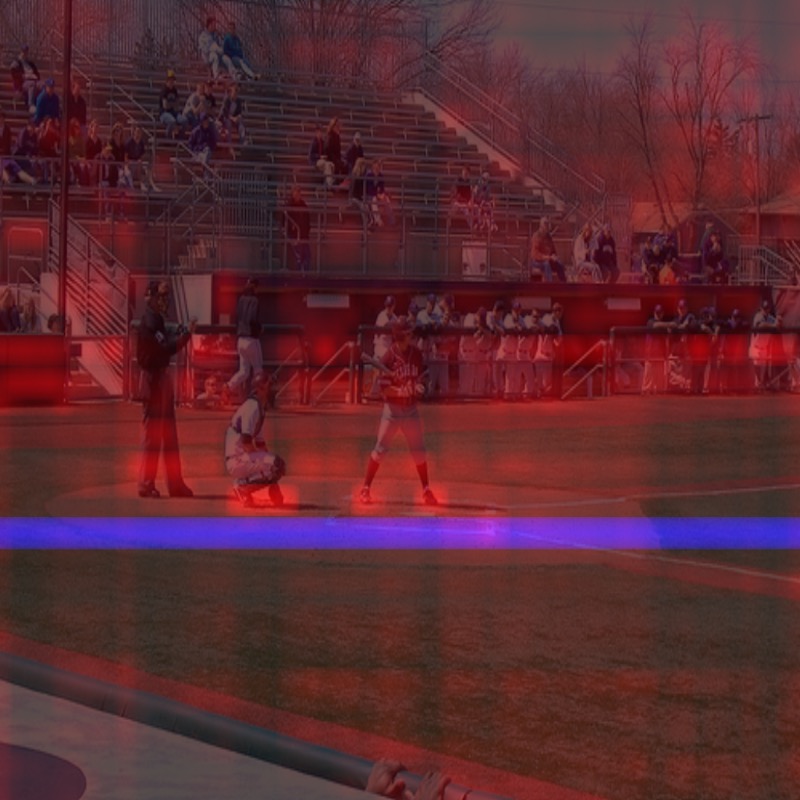} & \includegraphics[width=0.2000\textwidth]{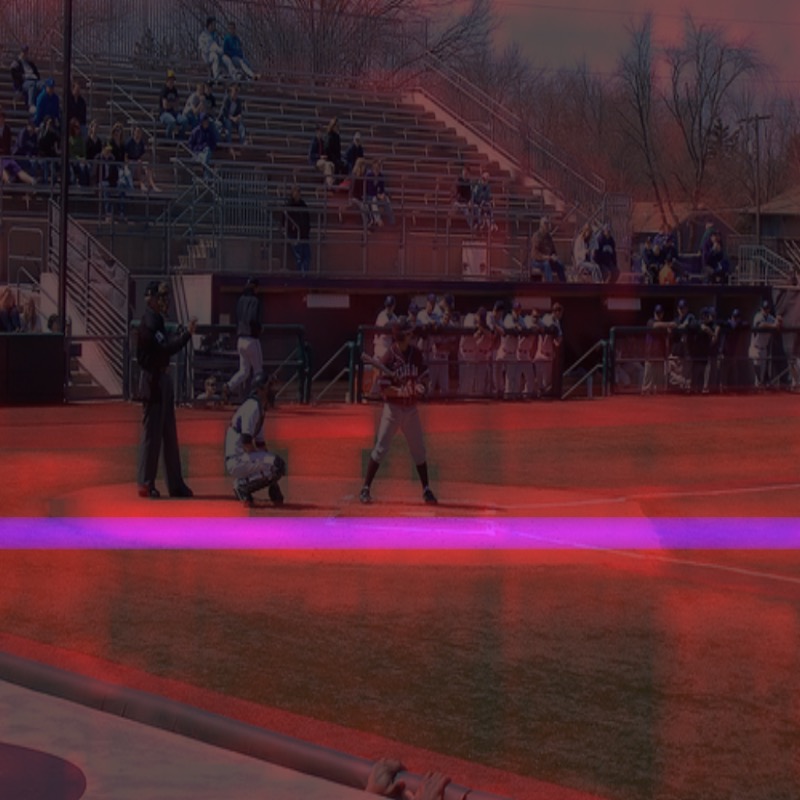} \\
    column head 5 & column head 6 & column head 7 & column head 8 \\
    \includegraphics[width=0.2000\textwidth]{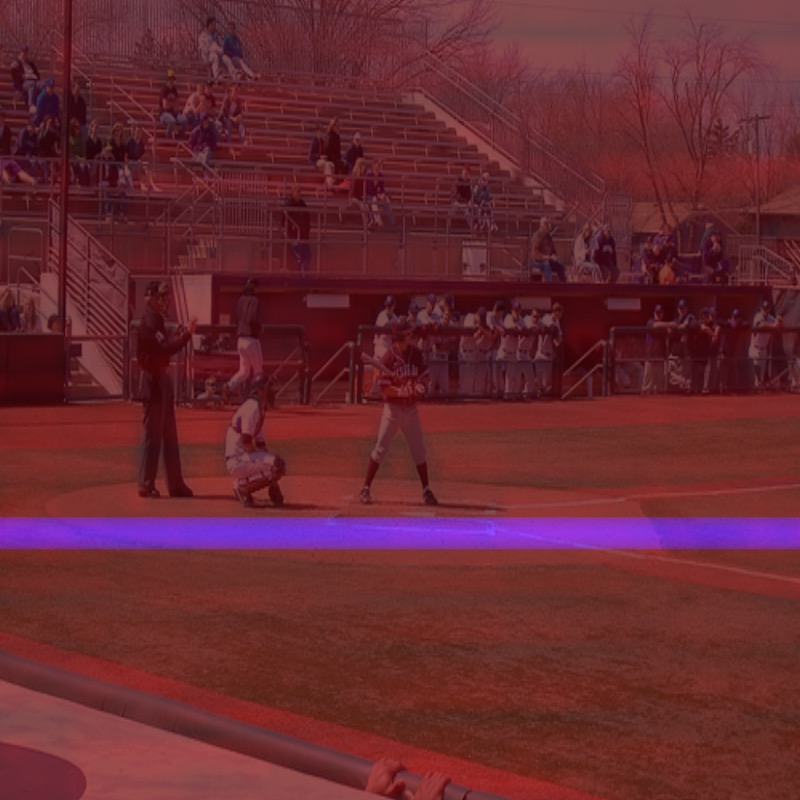} & \includegraphics[width=0.2000\textwidth]{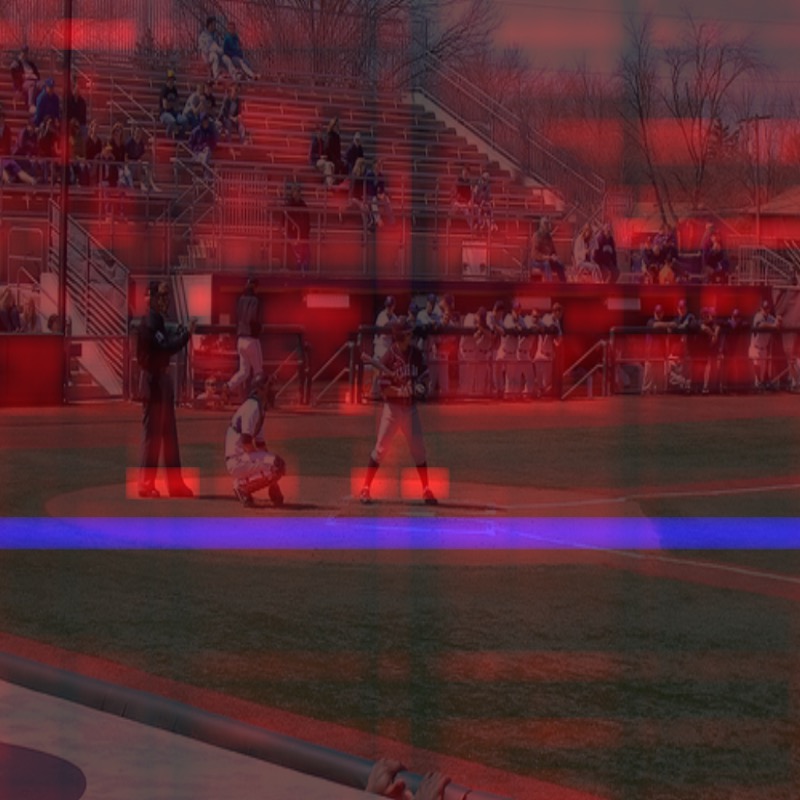} & \includegraphics[width=0.2000\textwidth]{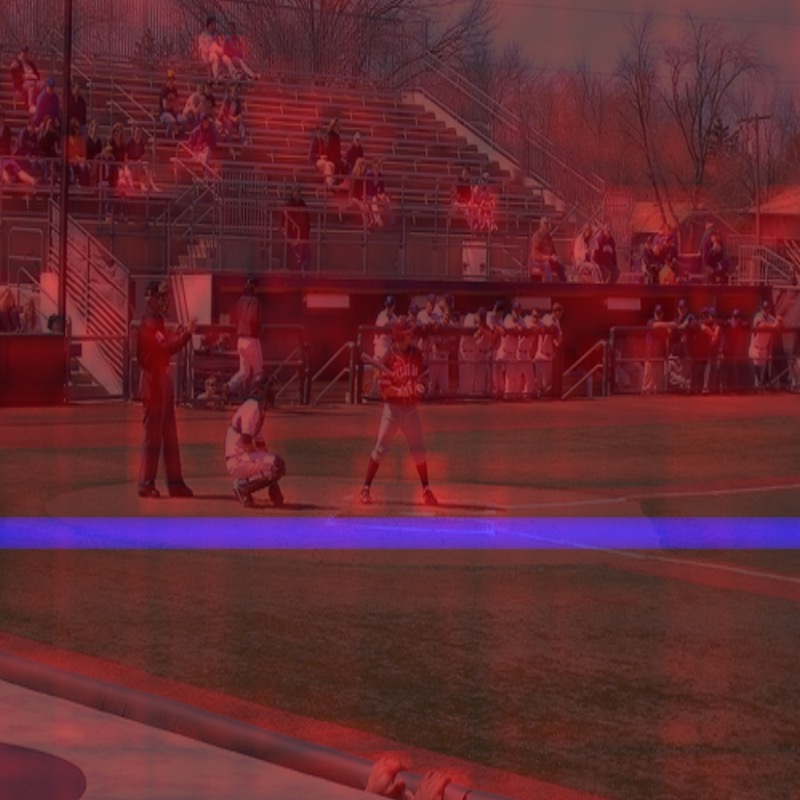} & \includegraphics[width=0.2000\textwidth]{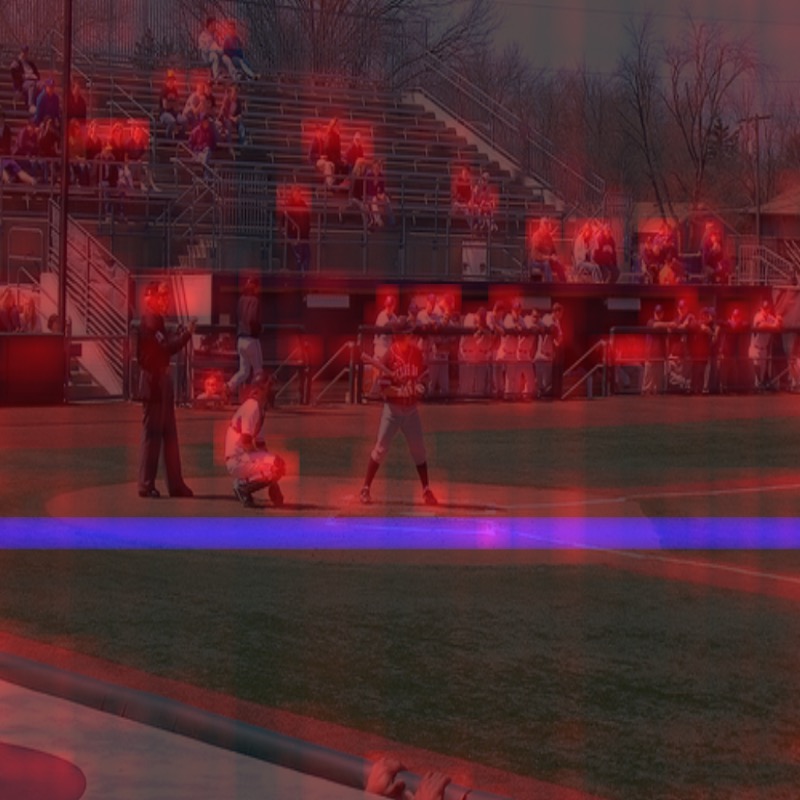} \\
    row head 1 & row head 2 & row head 3 & row head 4 \\
    \includegraphics[width=0.2000\textwidth]{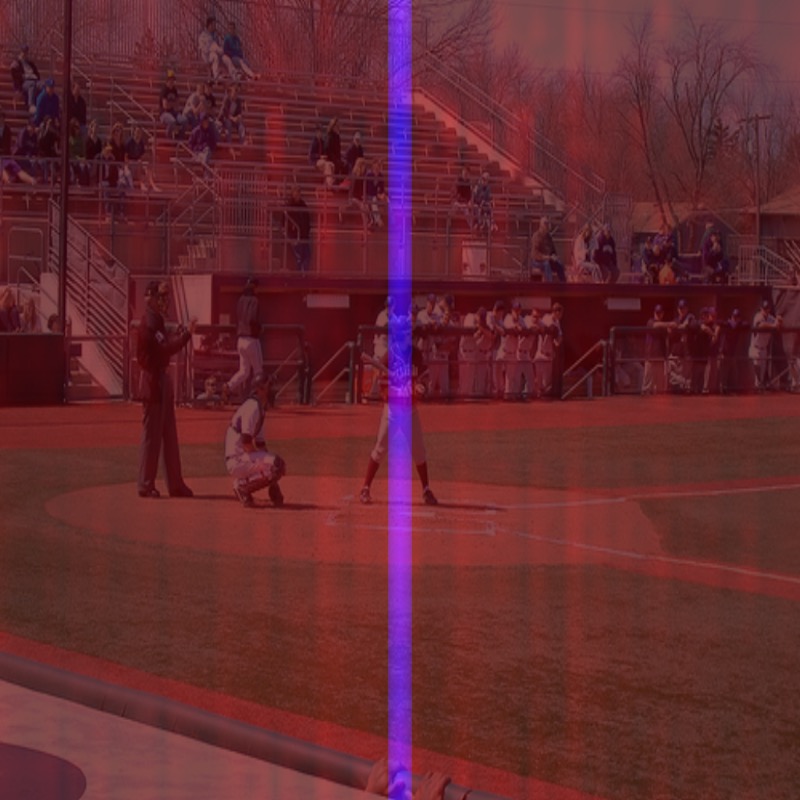} & \includegraphics[width=0.2000\textwidth]{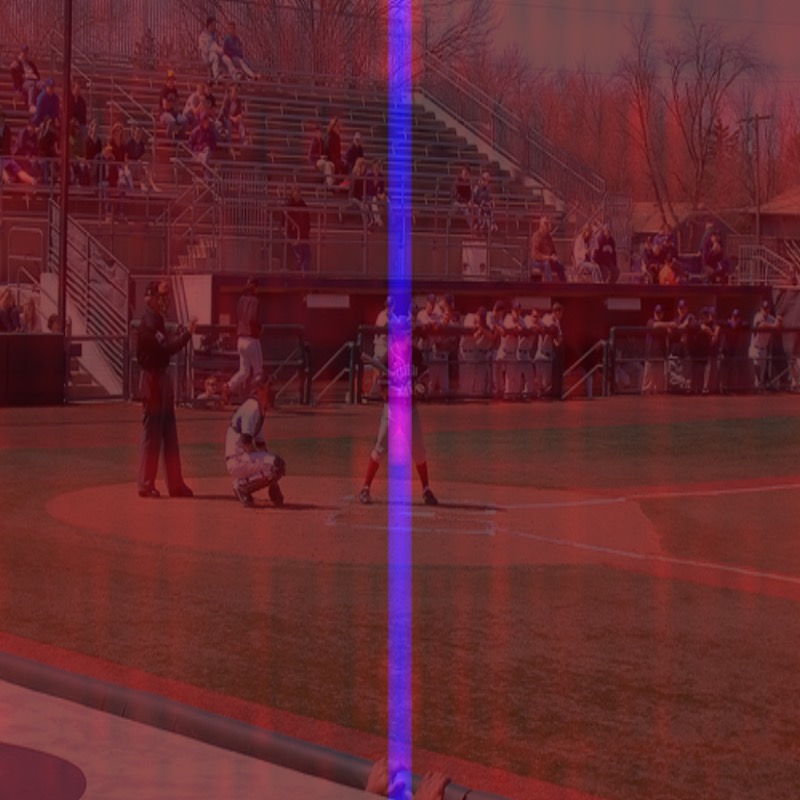} & \includegraphics[width=0.2000\textwidth]{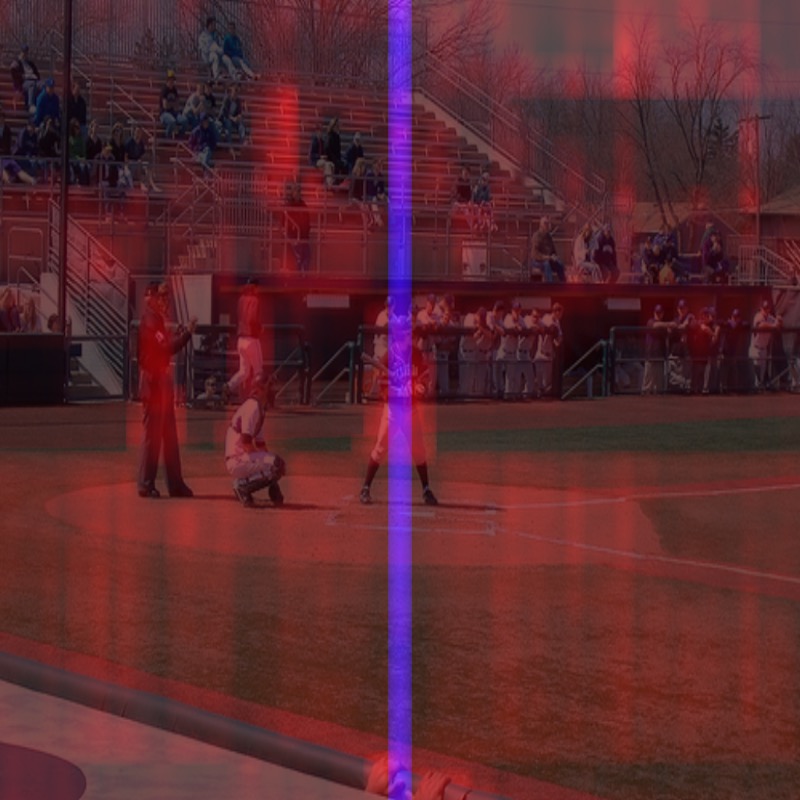} & \includegraphics[width=0.2000\textwidth]{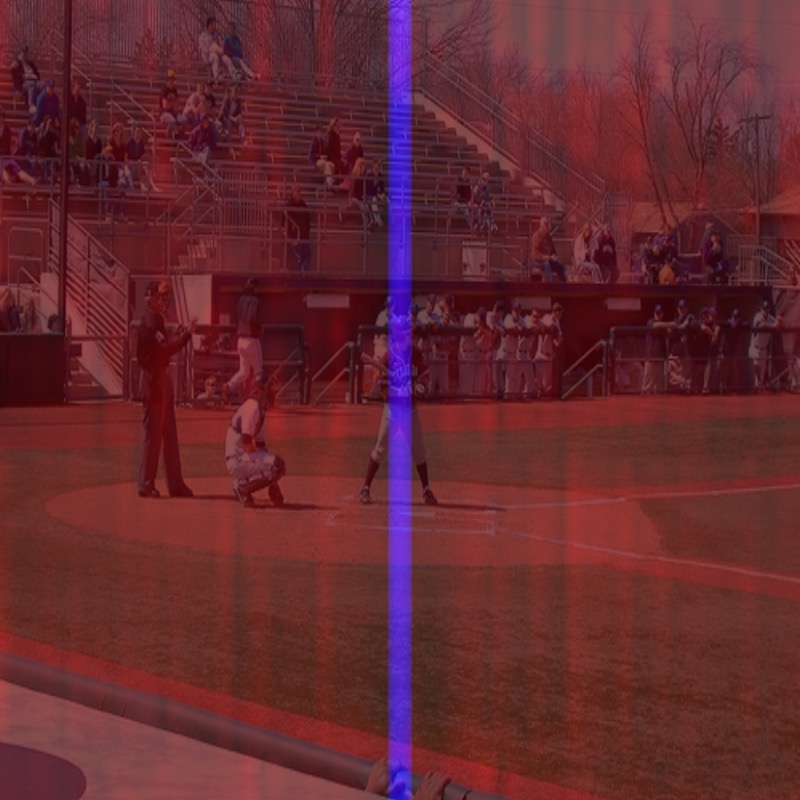} \\
    row head 5 & row head 6 & row head 7 & row head 8 \\
    \includegraphics[width=0.2000\textwidth]{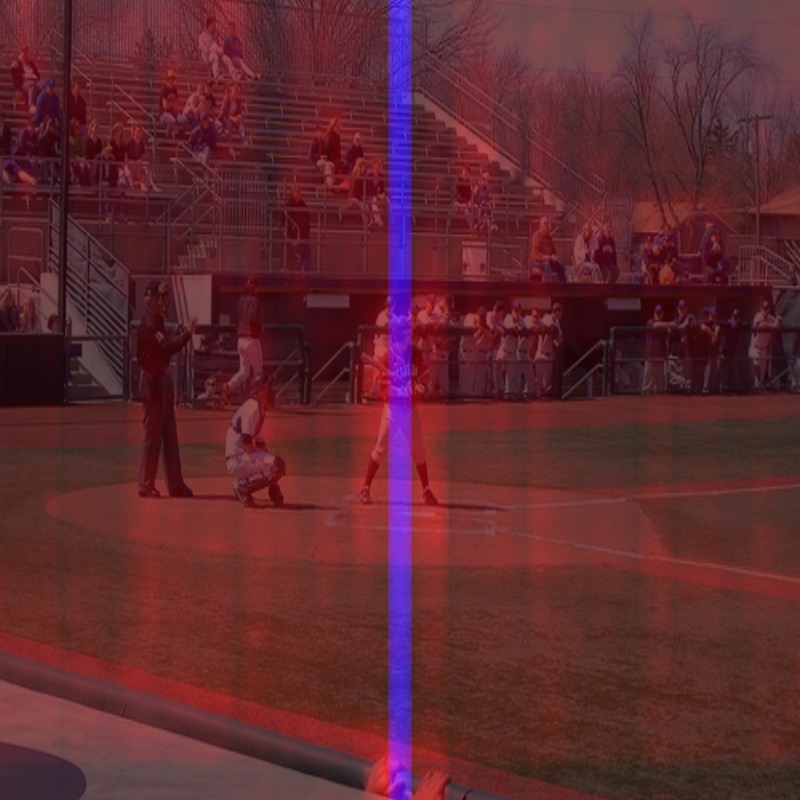} & \includegraphics[width=0.2000\textwidth]{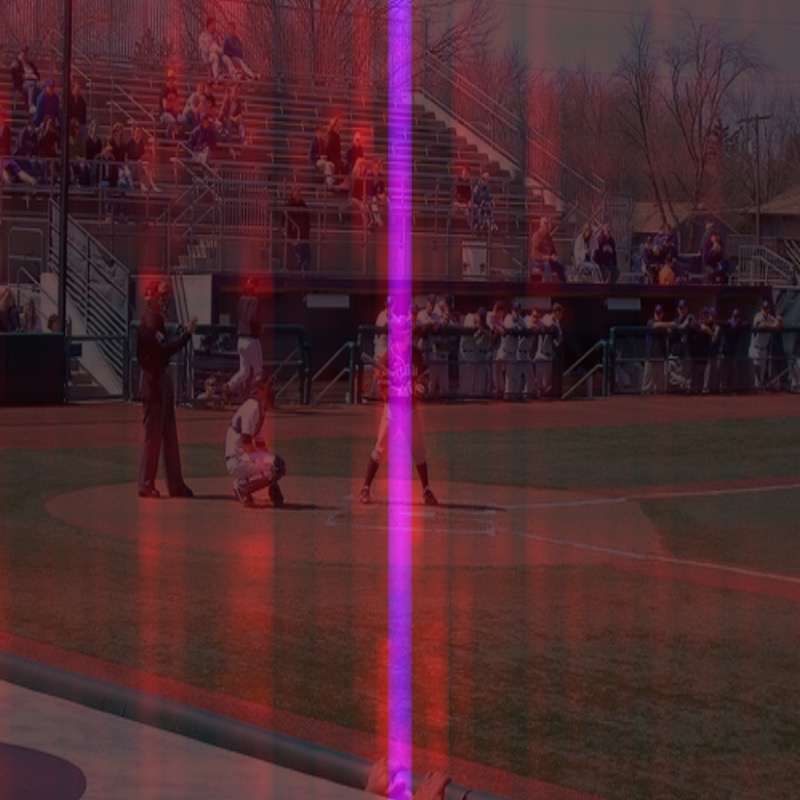} & \includegraphics[width=0.2000\textwidth]{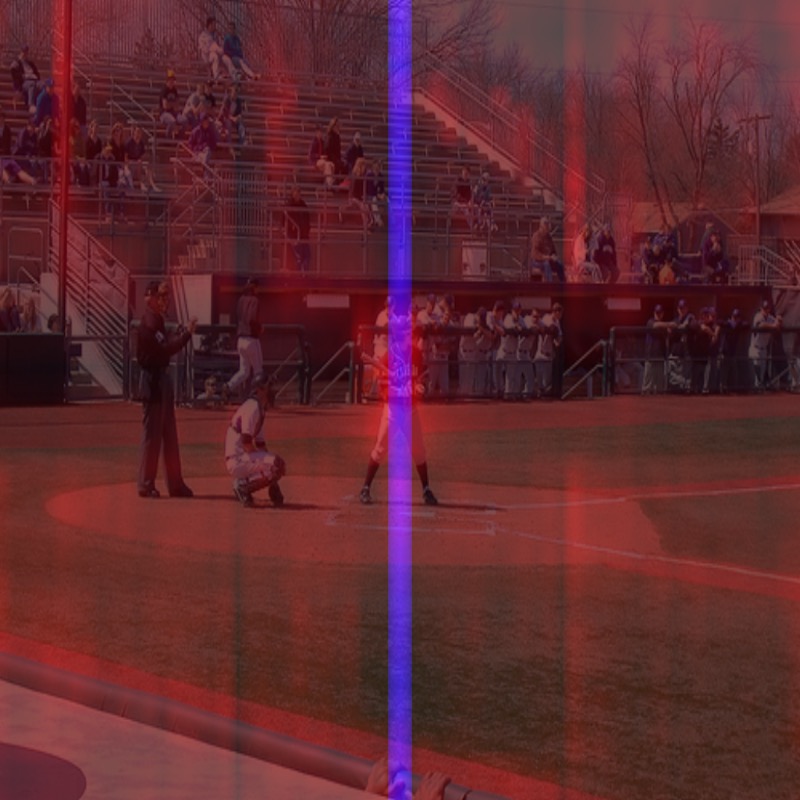} & \includegraphics[width=0.2000\textwidth]{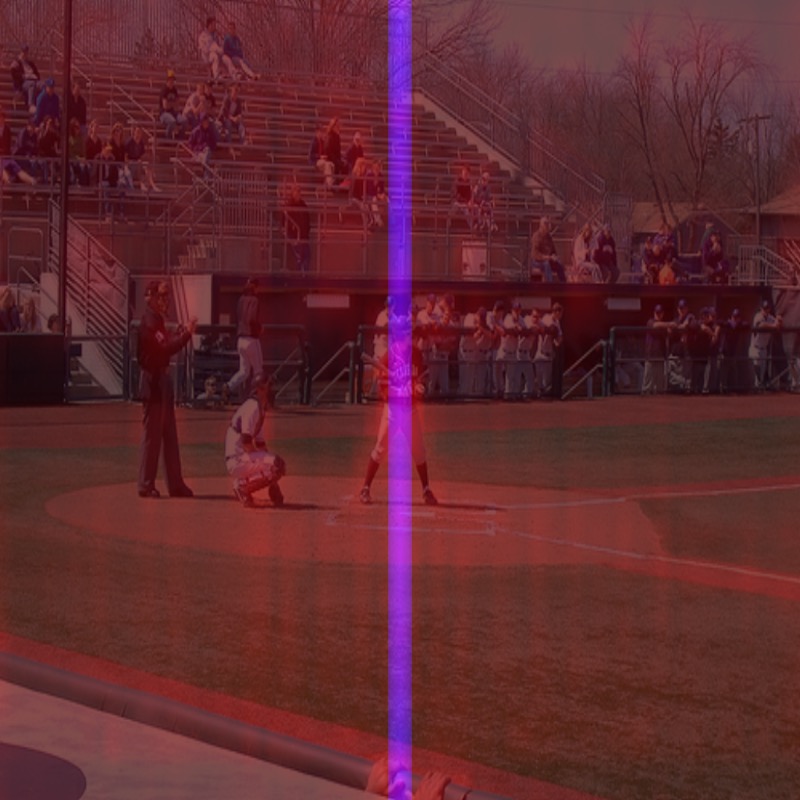} \\
    \end{tabular}
    \caption{Attention maps in block 2 of stage 3. We take a row of pixels, and visualize their column (height-axis) attention in all 8 heads. Then, we take a column, and visualize their row attention. Blue pixels are queries that we take, and red pixels indicate the corresponding attention weights. We notice that column head 1 corresponds to human heads, while column head 4 correlates with the field only. Row head 6 focuses on relatively local regions whereas column head 5 pools all over the whole image}
    \label{fig:attention1}
\end{figure}

\begin{figure}
    \centering
    \setlength\tabcolsep{3.12345pt}
    \begin{tabular}{cccc}
    \multicolumn{2}{c}{Original Image} & \multicolumn{2}{c}{Panoptic Prediction} \\
    \multicolumn{2}{c}{\includegraphics[width=0.35\textwidth]{figures_appendix/000044_image.jpg}} & \multicolumn{2}{c}{\includegraphics[width=0.35\textwidth]{figures_appendix/000044_panoptic_prediction.jpg}} \\
    column head 1 & column head 2 & column head 3 & column head 4 \\
    \includegraphics[width=0.2000\textwidth]{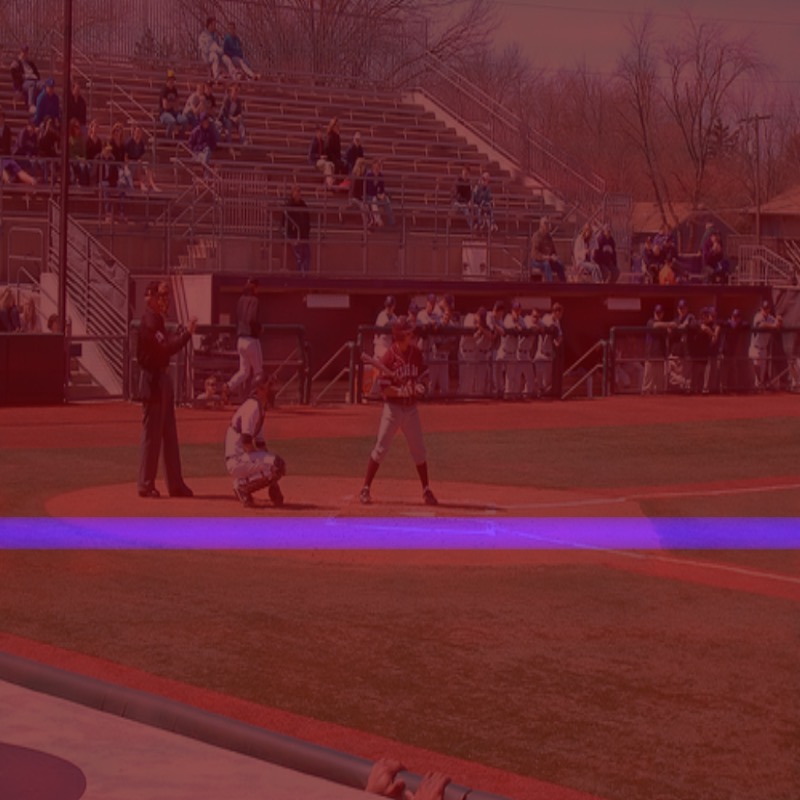} & \includegraphics[width=0.2000\textwidth]{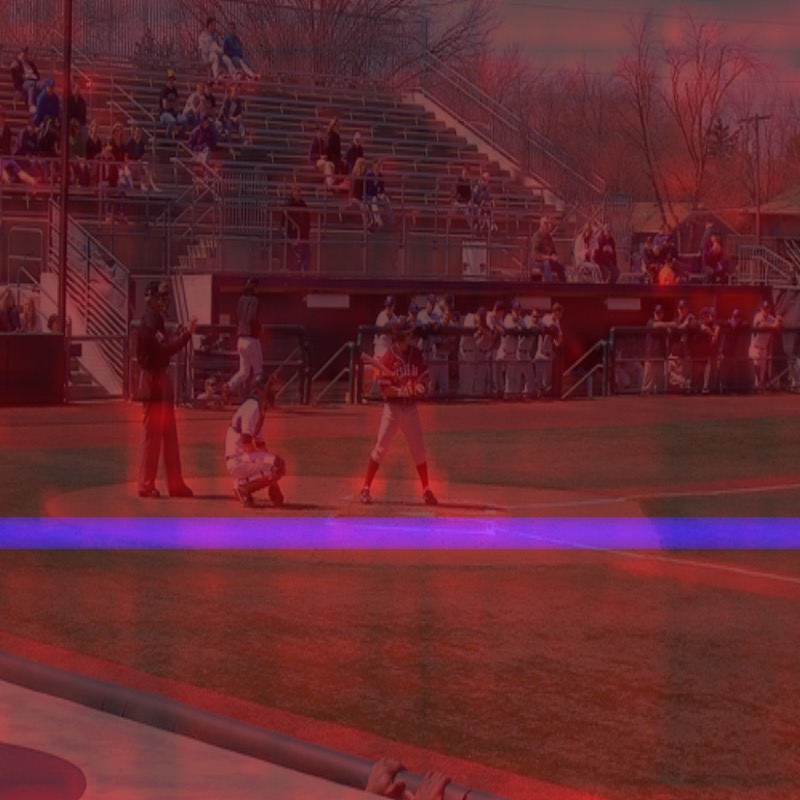} & \includegraphics[width=0.2000\textwidth]{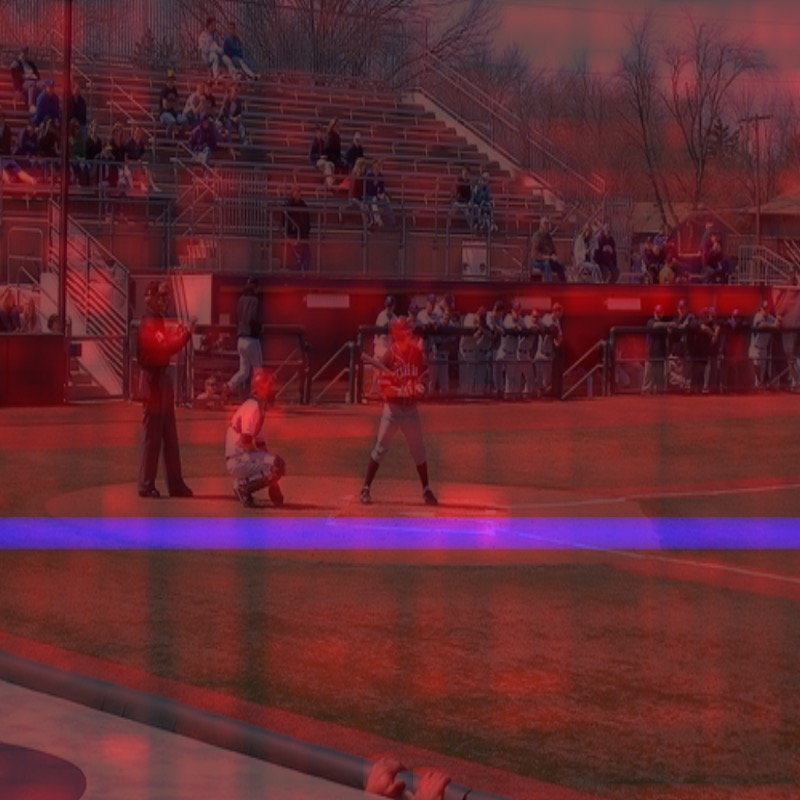} & \includegraphics[width=0.2000\textwidth]{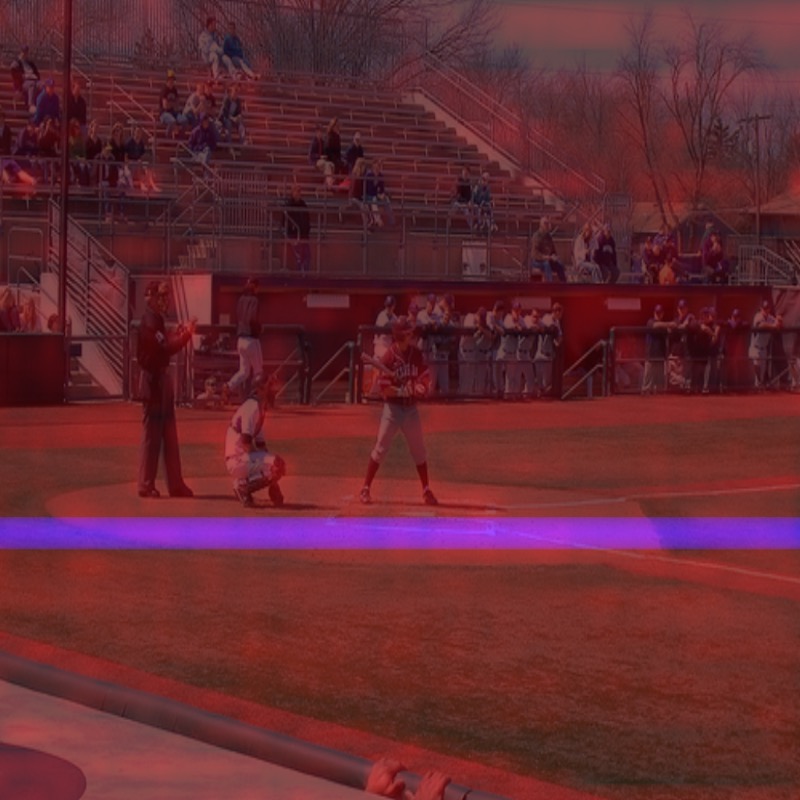} \\
    column head 5 & column head 6 & column head 7 & column head 8 \\
    \includegraphics[width=0.2000\textwidth]{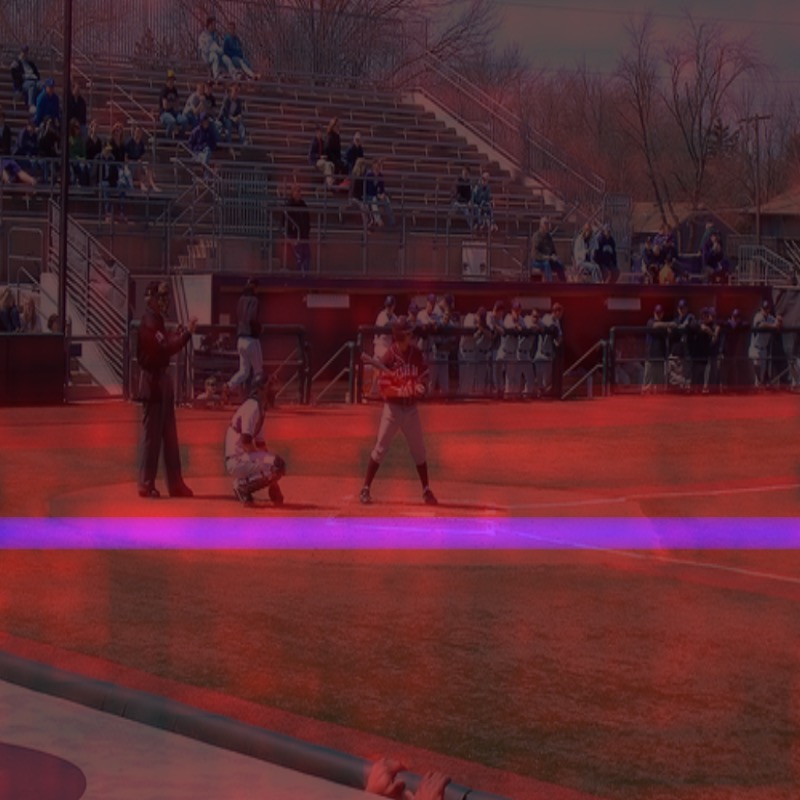} & \includegraphics[width=0.2000\textwidth]{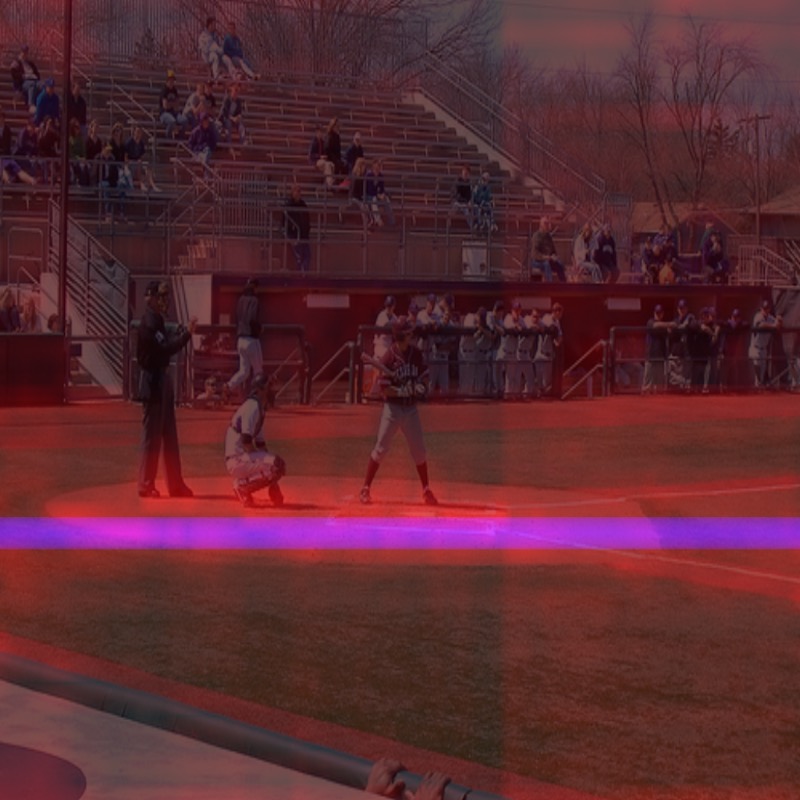} & \includegraphics[width=0.2000\textwidth]{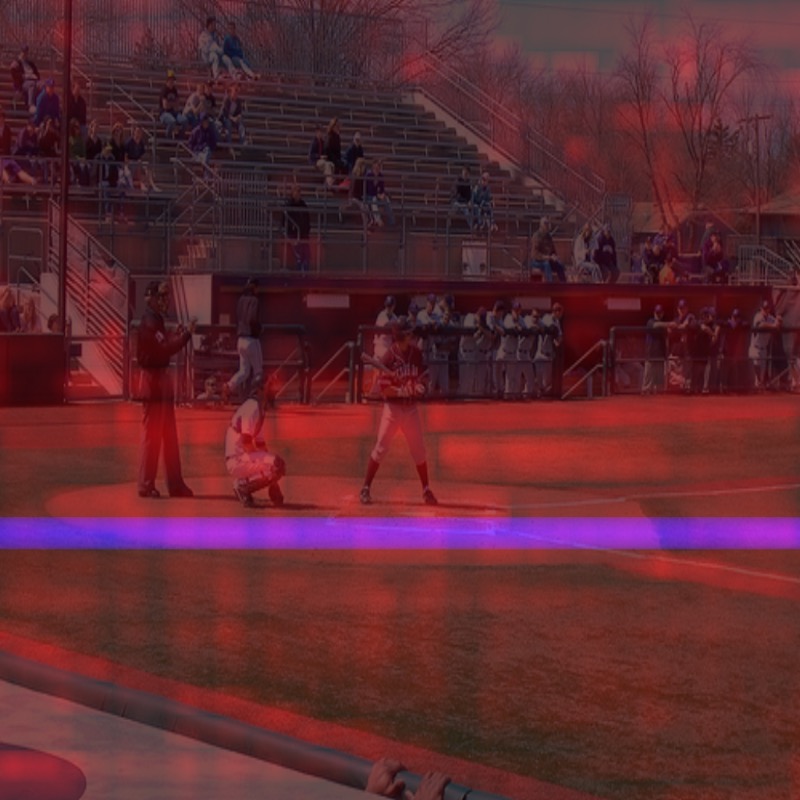} & \includegraphics[width=0.2000\textwidth]{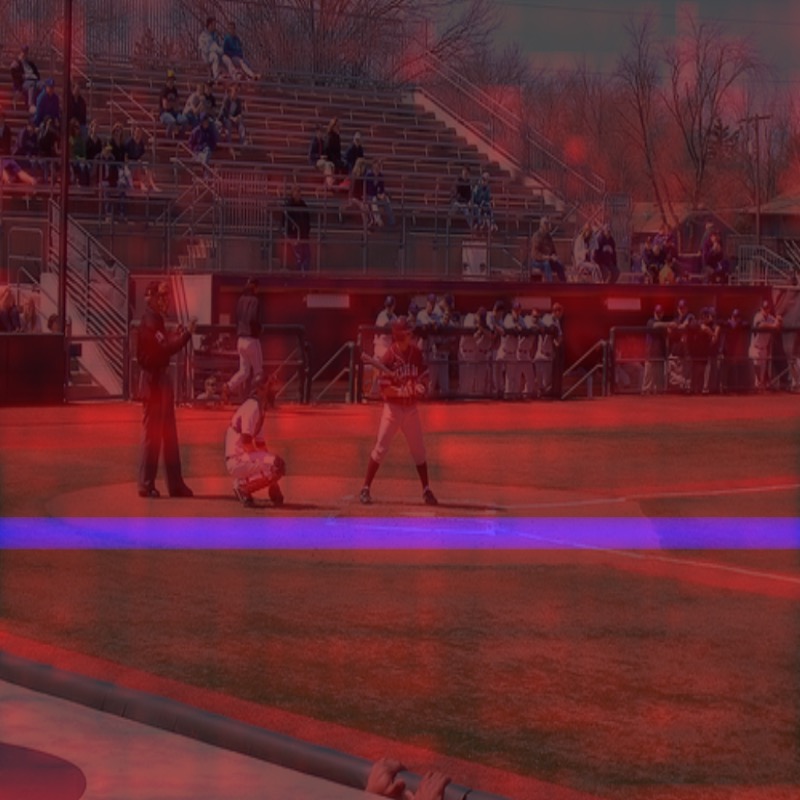} \\
    row head 1 & row head 2 & row head 3 & row head 4 \\
    \includegraphics[width=0.2000\textwidth]{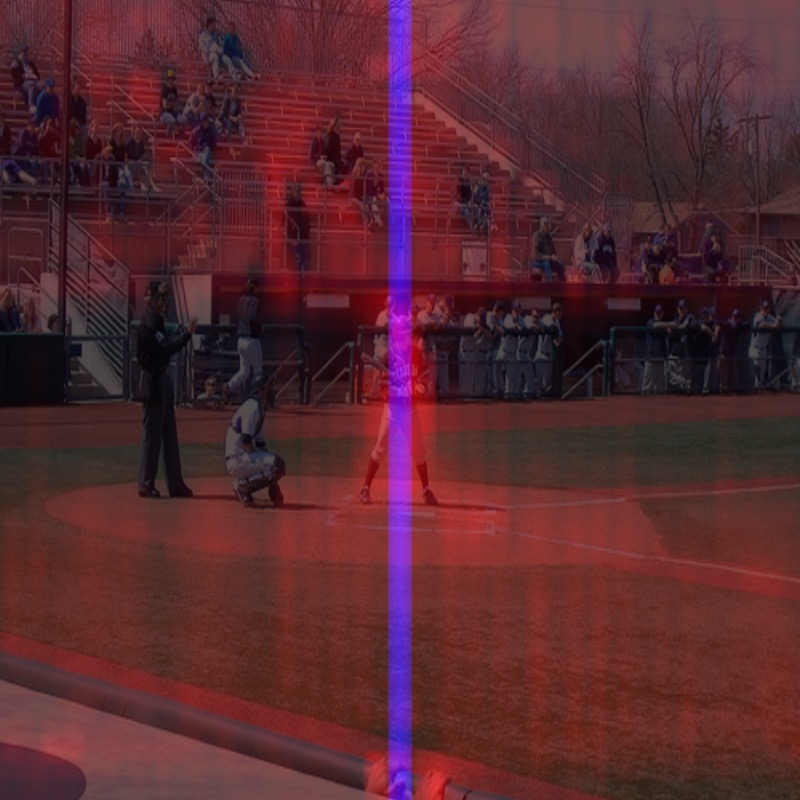} & \includegraphics[width=0.2000\textwidth]{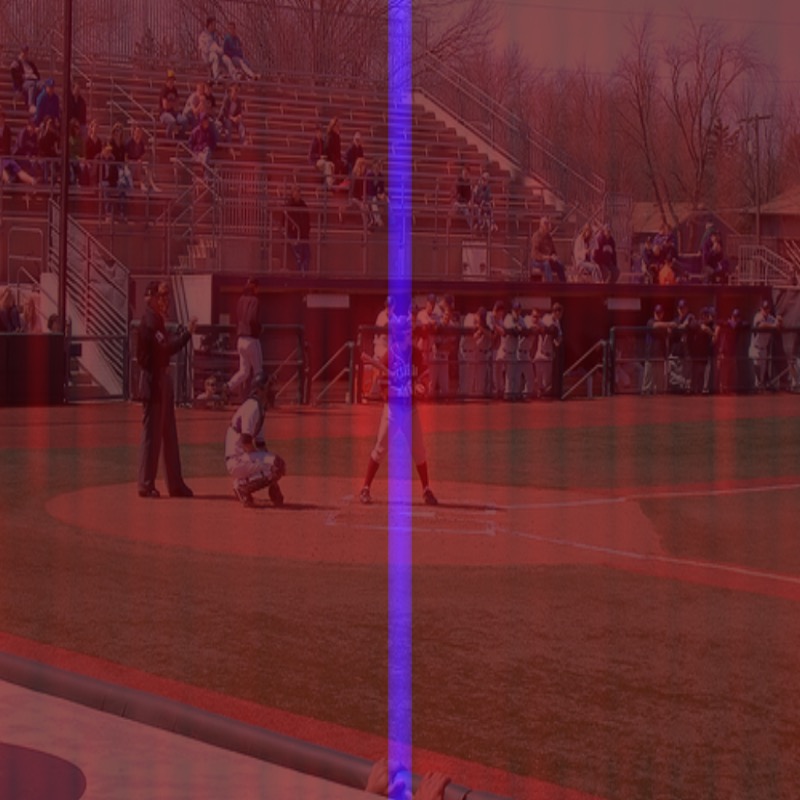} & \includegraphics[width=0.2000\textwidth]{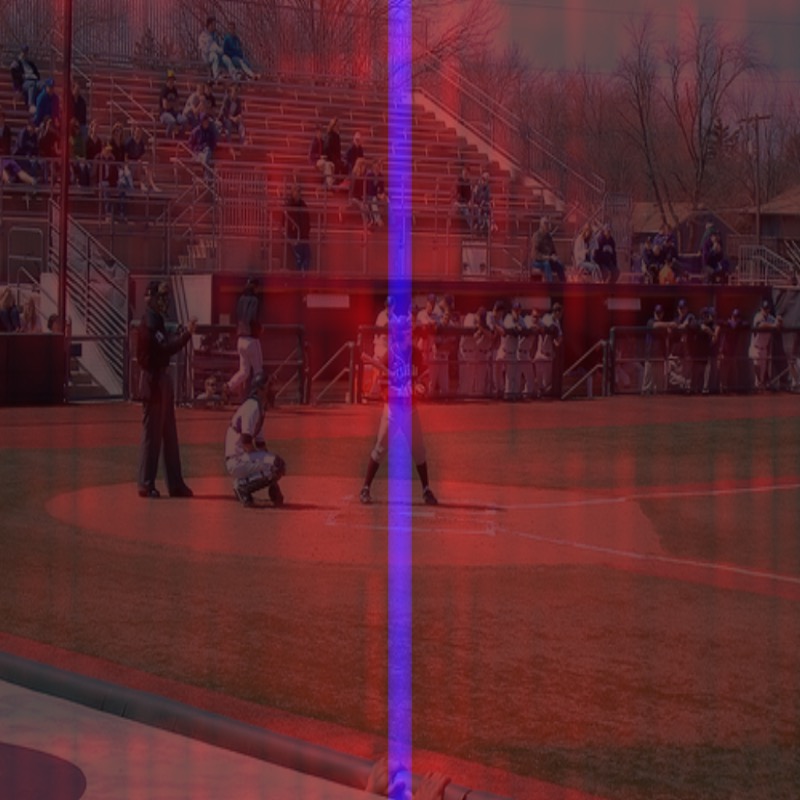} & \includegraphics[width=0.2000\textwidth]{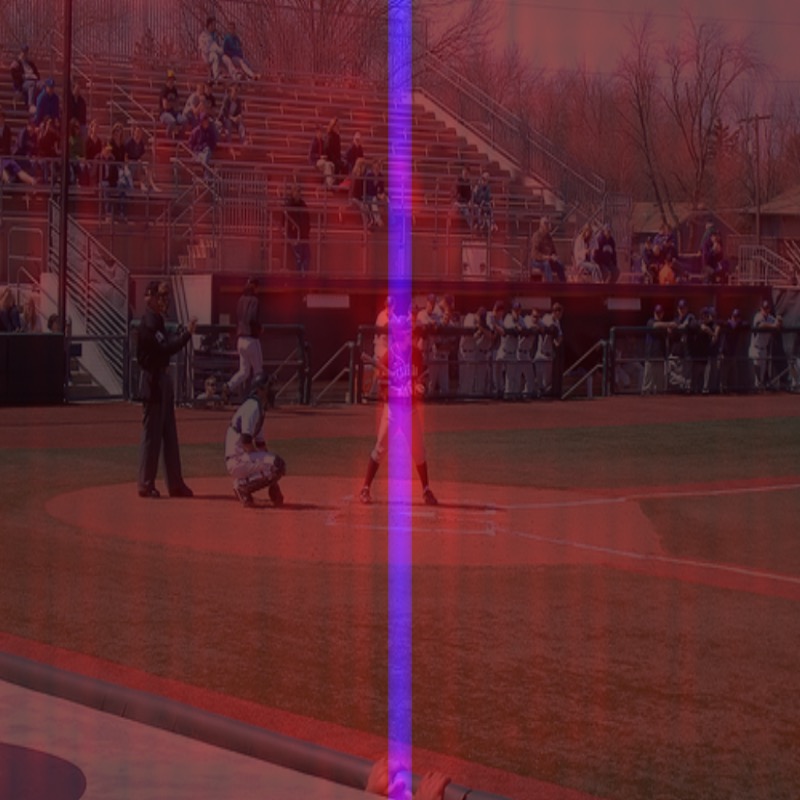} \\
    row head 5 & row head 6 & row head 7 & row head 8 \\
    \includegraphics[width=0.2000\textwidth]{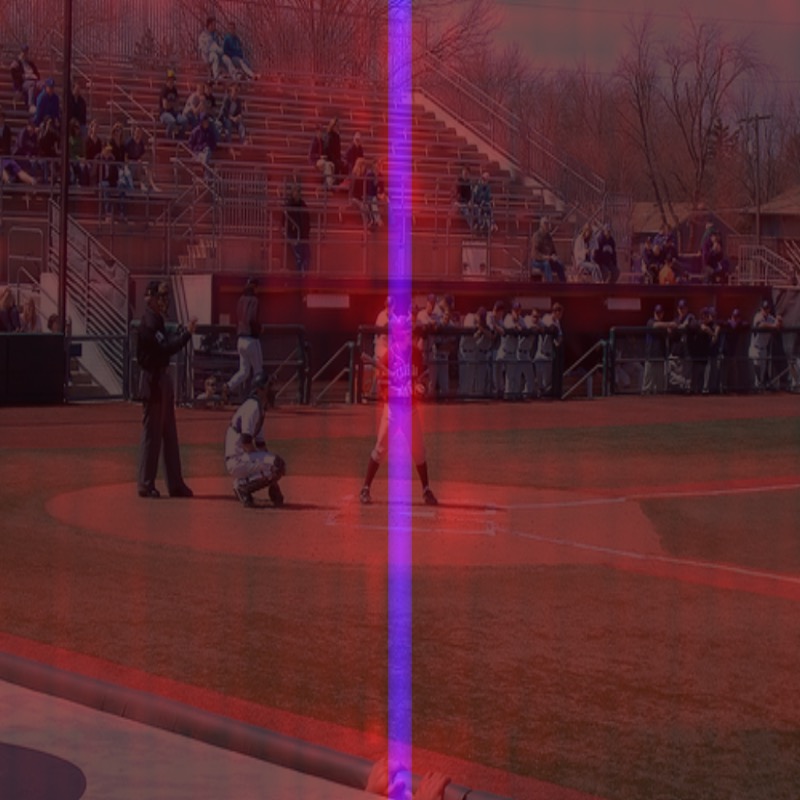} & \includegraphics[width=0.2000\textwidth]{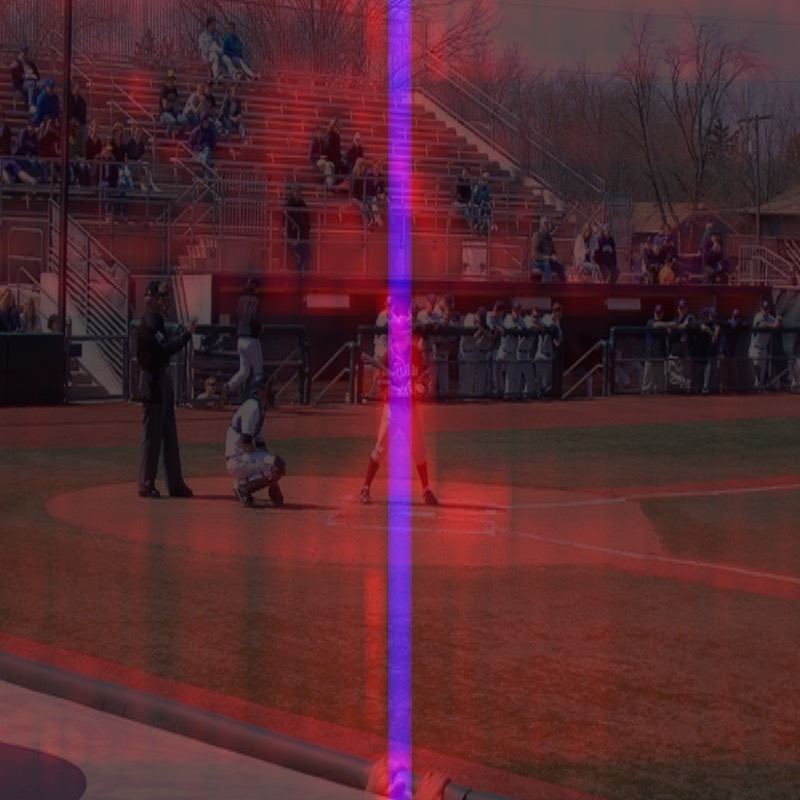} & \includegraphics[width=0.2000\textwidth]{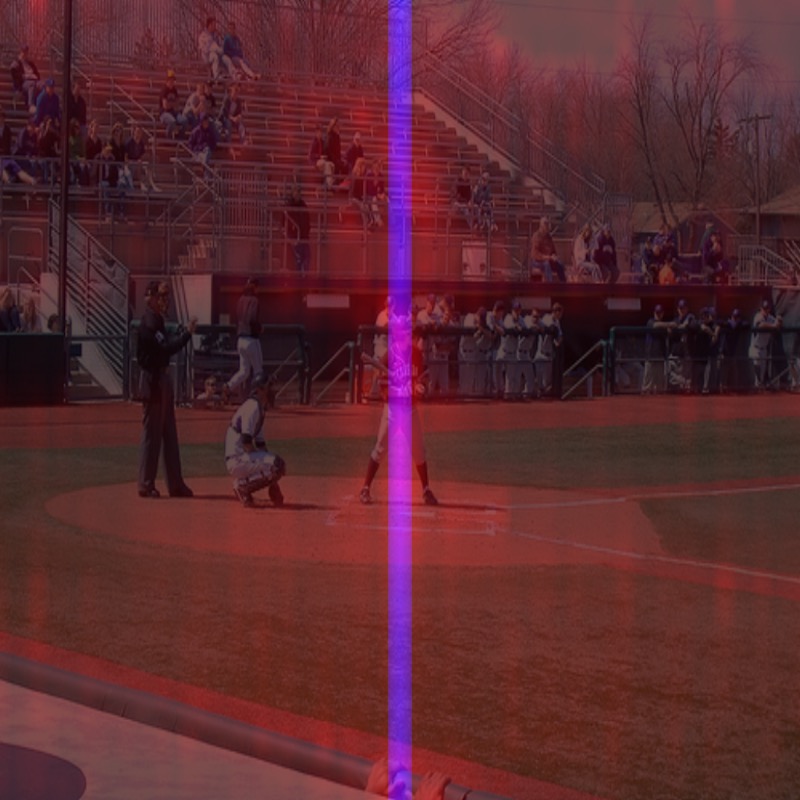} & \includegraphics[width=0.2000\textwidth]{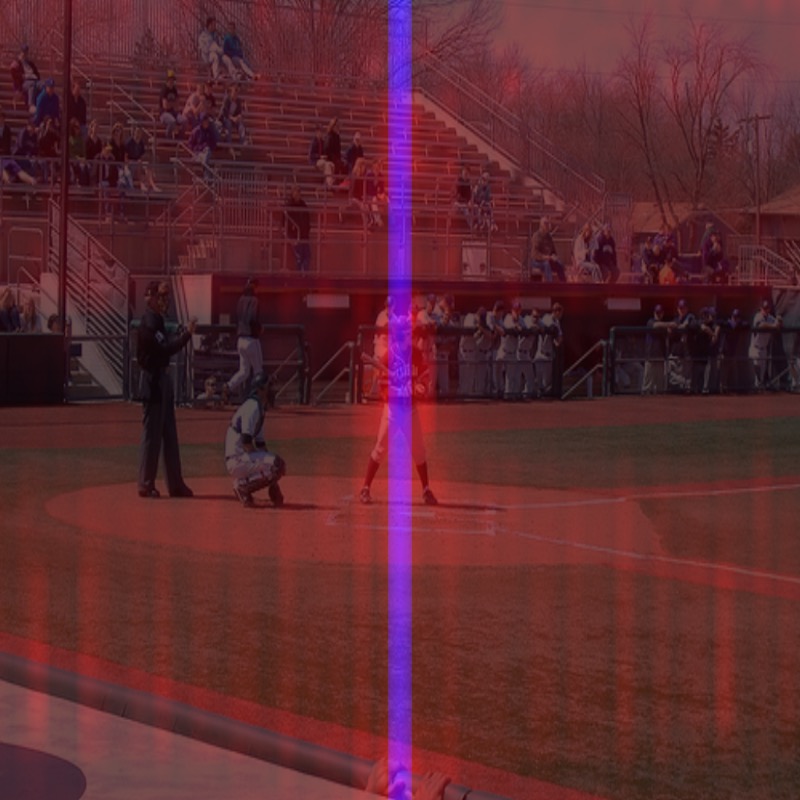} \\
    \end{tabular}
    \caption{Attention maps in block 3 of stage 4. They focus more on long range context than those in \figref{fig:attention1}, although all of them have a global receptive field}
    \label{fig:attention2}
\end{figure}

In \figref{fig:attention1} and \figref{fig:attention2}, we visualize the attention maps of our Axial-DeepLab-L on COCO val set. We visualize a low level block (stage 3 block 2) and a high level block (stage 4 block 3), which are respectively the first block and the last block with resolution 65$\times$65, in the setting of output stride 16. We notice that in our multi-head axial-attention, some heads learn to focus on local details while some others focus on long range context. Additionally, we find that some heads are able to capture positional information and some others learn to correlate with semantic concepts

\begin{figure}[t]
    \centering
    \includegraphics[width=0.7\textwidth]{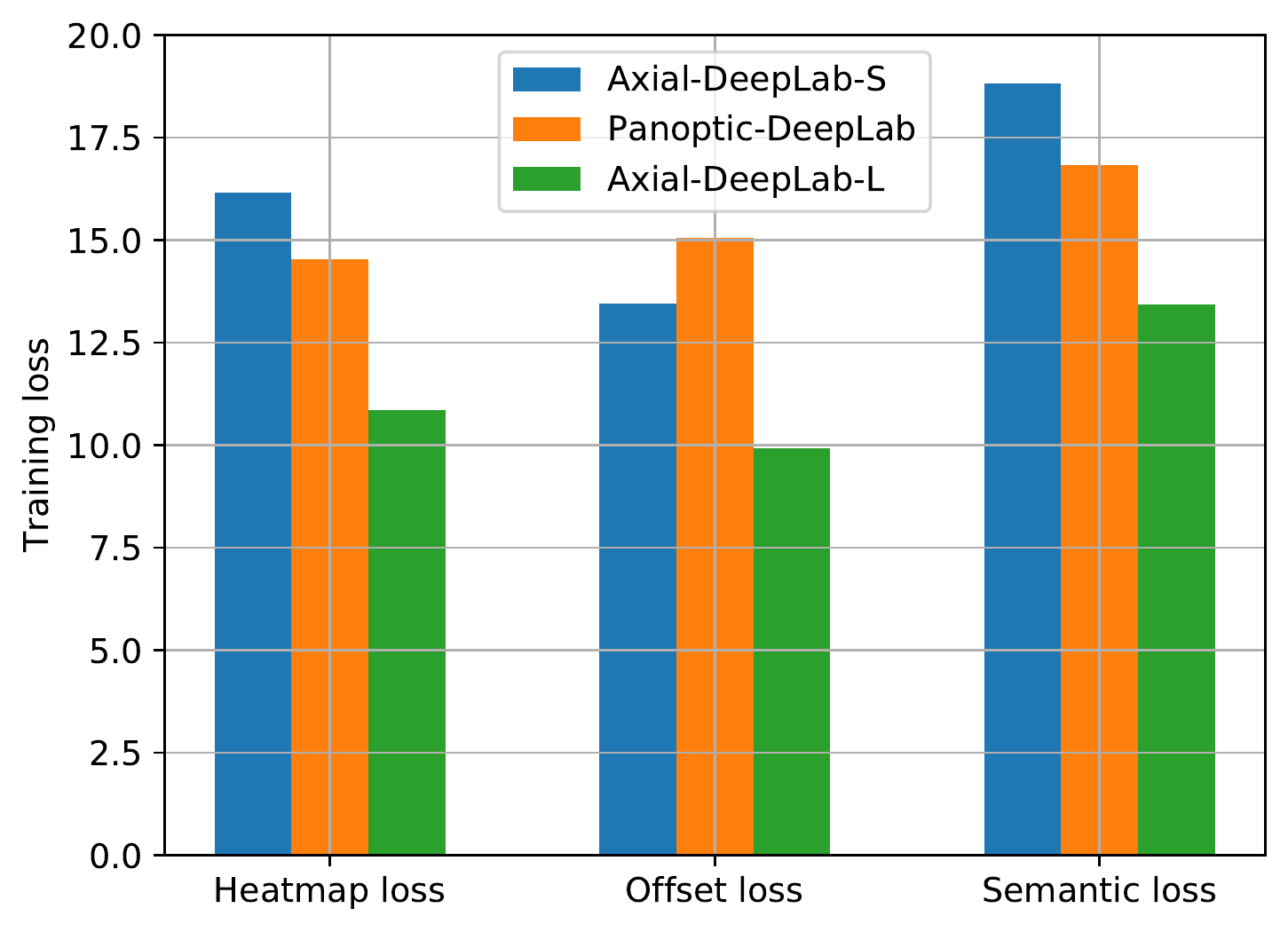}
    \caption{Training loss on COCO. Equipped with position-sensitive axial-attention, our Axial-DeepLab fits data distribution better than Panoptic-DeepLab~\cite{cheng2019panoptic}, especially on the task of predicting the offset to the object center, which requires precise and long range positional information}
    \label{fig:loss}
\end{figure}

In \figref{fig:loss}, we compare Axial-DeepLab with Panoptic-DeepLab~\cite{cheng2019panoptic}, in terms of the three training loss functions, defined in Panoptic-DeepLab~\cite{cheng2019panoptic}. We observe that Axial-DeepLab is able to fit data better, especially on the offset prediction task. This also demonstrates the effectiveness of our position-sensitive attention design, and the long range modeling ability of axial-attention.

\section{Raw Data}
In companion to \figref{fig:params_flops} of the main paper where we compare parameters and M-Adds against accuracy on ImageNet classification, we also show the performance of our models in \tabref{tab:imagenet2}.

\begin{table}[!t]
\setlength{\tabcolsep}{1.0em}
    \centering
    \caption{ImageNet validation set results. \textbf{Width:} the width multiplier that scales the models up. \textbf{Full:} Stand-alone self-attention models without spatial convolutions}
    \label{tab:imagenet2}
    \scalebox{0.9}{
    \begin{tabular}{l|c|c|c|c|c}
        \toprule[0.2em]
        Method & Width & Full & Params & M-Adds & Top-1 \\
        \toprule[0.2em]
        Conv-Stem + PS-Attention & 0.5  & & 5.1M & 1.2B & 75.5 \\
        Conv-Stem + PS-Attention & 0.75 & & 10.5M & 2.3B & 77.4 \\
        Conv-Stem + PS-Attention & 1.0  & & 18.0M & 3.7B & 78.1 \\
        Conv-Stem + PS-Attention & 1.25 & & 27.5M & 5.6B & 78.5 \\
        Conv-Stem + PS-Attention & 1.5  & & 39.0M & 7.8B & 79.0 \\
        \midrule
        Conv-Stem + Axial-Attention & 0.375 & & 7.4M & 1.8B & 76.4 \\
        Conv-Stem + Axial-Attention & 0.5   & & 12.4M & 2.8B & 77.5 \\
        Conv-Stem + Axial-Attention & 0.75  & & 26.4M & 5.7B & 78.6 \\
        Conv-Stem + Axial-Attention & 1.0.  & & 45.6M & 9.6B & 79.0 \\
        \midrule
        Full Axial-Attention & 0.5  & \ding{51} & \textbf{12.5M} & \textbf{3.3B} & \textbf{78.1} \\
        Full Axial-Attention & 0.75 & \ding{51} & \textbf{26.5M} & \textbf{6.8B} & \textbf{79.2} \\
        Full Axial-Attention & 1.0  & \ding{51} & \textbf{45.8M} & \textbf{11.6B} & \textbf{79.3} \\
        \bottomrule[0.1em]
    \end{tabular}
    }
\end{table}

\begin{figure}[!t]
    \centering
    \includegraphics[width=0.66\textwidth]{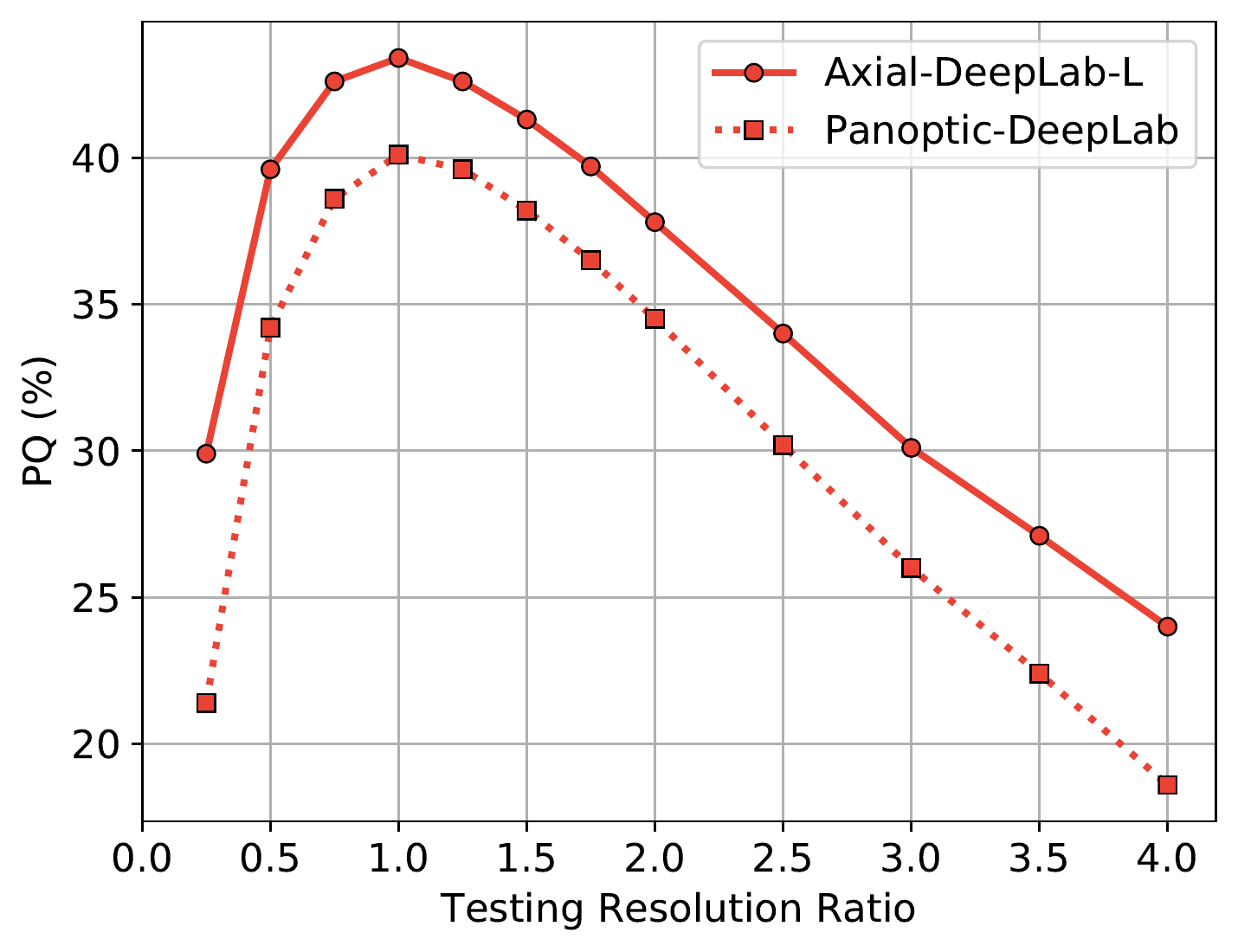}
    \caption{Scale stress test on COCO val set}
    \label{fig:scale2}
\end{figure}

In companion to \figref{fig:scale} of the main paper where we demonstrate the relative improvements of Axial-DeepLab-L over Panoptic-DeepLab (Xception-71) in our scale stress test on COCO, we also show the raw performance of both models in \figref{fig:scale2}.

\end{subappendices}

%
%
\FloatBarrier
\bibliographystyle{splncs04}
\bibliography{99_references}
\end{document}